\documentclass[preprint,12pt]{elsarticle}

\usepackage{graphicx}
\usepackage{caption}
\usepackage{subcaption}
\usepackage[table,xcdraw]{xcolor}
\usepackage{url}
\PassOptionsToPackage{hyphens}{url}
\usepackage{hyperref}
\urldef{\giturl}\url{https://github.com/fmohr/lcdb/blob/ce96fa3768da94d222644883a11403119844f241/publications/2024-neurocom/multi-fidelity-hpo/README.md}
\usepackage{booktabs}
\usepackage{longtable}
\usepackage{comment}
\usepackage{placeins}

\usepackage{amssymb}
\usepackage{amsmath}
\usepackage[disable]{todonotes}

\newcommand{\tomhides}[1]{}

\definecolor{blue}{HTML}{1f77b4}
\definecolor{orange}{HTML}{ff7f0e}
\definecolor{green}{HTML}{2ca02c}
\definecolor{red}{HTML}{d62728}
\definecolor{purple}{HTML}{9467bd}
\definecolor{darkblue}{HTML}{000055}
\definecolor{darkred}{HTML}{880000}
\definecolor{blueplot}{HTML}{08cff9}
\definecolor{yellowplot}{HTML}{ffc300}

\journal{Neurocomputing}

\begin{document}

\begin{frontmatter}



\title{The Unreasonable Effectiveness Of Early Discarding After One Epoch In Neural Network Hyperparameter Optimization}

%\title{Efficiency in Simplicity: The Unreasonable Effectiveness of Single-Epoch Evaluation in Deep Neural Network Hyperparameter Optimization.}


% COMMENT: Authors add your name + affiliation here, we can discuss about the order later
\author[anl,upsaclay]{Romain Egele}
\author[unisabana]{Felix Mohr}
\author[tud]{Tom Viering}
\author[ornl]{Prasanna Balaprakash}

\affiliation[anl]{organization={Argonne National Laboratory},%Department and Organization
            country={USA}}
\affiliation[upsaclay]{organization={Université Paris-Saclay},%Department and Organization
            country={France}}
\affiliation[unisabana]{organization={Universidad de La Sabana},%Department and Organization
            country={Colombia}}
\affiliation[tud]{organization={Delft University of Technology},%Department and Organization
            country={Netherlands}}
\affiliation[ornl]{organization={Oak Ridge National Laboratory},
            country={USA}}

\begin{abstract}
To reach high performance with deep learning, hyperparameter optimization (HPO) is essential.
This process is usually time-consuming due to costly evaluations of neural networks.
Early discarding techniques limit the resources granted to unpromising candidates by observing the empirical learning curves and canceling neural network training as soon as the lack of competitiveness of a candidate becomes evident.
Despite two decades of research, little is understood about the trade-off between the aggressiveness of discarding and the loss of predictive performance.
Our paper studies this trade-off for several commonly used discarding techniques such as successive halving and learning curve extrapolation.
Our surprising finding is that these commonly used techniques offer minimal to no added value compared to the simple strategy of discarding after a constant number of epochs of training.
The chosen number of epochs mostly depends on the available compute budget.
We call this approach $i$-Epoch ($i$ being the constant number of epochs with which neural networks are trained) and suggest to assess the quality of early discarding techniques by comparing how their Pareto-Front (in consumed training epochs and predictive performance) complement the Pareto-Front of $i$-Epoch.
\end{abstract}


% \isabelle{Add legend to figure: what should we see/understand?}
% \centering
% \includegraphics{figures/lcdb2/554/curves_log_test.jpg}
% hello

% \begin{highlights}
% \item Early discarding deep neural networks after one epoch of training often achieves competitive predictive accuracy in hyperparameter optimization while using significantly less compute.
% \item Early discarding deep neural networks after a constant number of training epochs offers all the necessary trade-offs between predictive accuracy and necessary computations in hyperparameter optimization.
% \end{highlights}


\end{frontmatter}


\section{Introduction}

%%% General context of the work
Optimizing the configuration of a deep learning pipeline is a complex task that involves properly configuring the data preprocessing, training algorithm, and neural architecture.
A configuration is a specification of so-called hyperparameters~\cite{yu_hyper-parameter_2020}, which control the behavior of pipeline elements and hence can greatly influence its final predictive performance.
The objective is to identify the configuration of hyperparameters that achieves the best predictive performance, usually referred to as hyperparameter optimization (HPO).

%%% Narrower research area and statement of its importance
As HPO is often done from a black-box optimization point of view, that is by observation of input configuration and output performance, a major challenge is the required computation to evaluate candidate hyperparameters by training deep neural networks.
This greatly limits the number of testable hyperparameter configurations within a practical time frame.
This is why multi-fidelity hyperparameter optimization with early discarding was proposed to switch the black-box problem to a ``gray-box'' optimization problem by observing the intermediate training performance of neural networks and using it as an estimate of the final performance. 
Such estimates can in principle be obtained at a computationally cheaper training stage and therefore save overall computation. 
In deep neural networks, the training epochs are usually used to perform early discarding. An epoch usually refers to making a full pass over the training data.
The predictive performance versus the number of epochs is also known as a ``learning curve'' \cite{viering2022shape,mohr_learning_2022}.

HPO with early discarding trades-off computation with quality of extrapolated performance.
For example, if the neural network is trained for a few epochs, it can save computation but it also means we have little (noisy) training information and therefore increase the chances of mistaking the extrapolation.
It is important to note that extrapolated performance is not always absolute but it can also be relative to other candidates such as by predicting a ranking.

%%% Announcement of principal findings
A shortcoming of the HPO early discarding literature is the multi-objective ((1) predictive performance, (2) overall computation) optimization viewpoint that such techniques are trying to solve. 
Therefore, experimental evaluations lack comparison to proper baselines and sometimes present over-optimistic results.
For example, it is common to compare early discarding techniques with complete training discarding~\cite{falkner_bohb_2018} and, only rare works consider the baseline performance, which minimizes computation by stopping the training after a single epoch~\cite{egele2023one,bohdal2023pasha} during HPO and possibly selecting from the top-$k$ models after further training. We call this baseline ``1-Epoch'' or more generally $i$-Epoch when the training is stopped after epoch $i$.

In this work, we evaluate the computation optimal policy 1-Epoch and show its surprising effectiveness in detecting top-ranked hyperparameter configurations.
In addition, we look at the set of trade-offs between computation and predictive performance offered by different early discarding methods among which is the $i$-Epoch baseline. 
We do this by spanning different levels of early discarding aggressiveness of each technique.
Being more aggressive (i.e., stopping training earlier) reduces computation but also generally sacrifices predictive performance. 
Therefore, we evaluate the multi-objective optimal frontier, also known as the Pareto-front, achieved by the different early discarding techniques. Ideally, varying the aggressiveness parameters of the different techniques, leads to a large Pareto-front, offering different trade-offs between aggressiveness (training epochs used) and predictive performance.  

To simplify our experiments and avoid confounding factors, we do not use advanced HPO solvers but instead perform a random sampling of hyperparameter configurations, for which we can compare several early discarding techniques. We compare $i$ -Epoch to asynchronous successive halving (SHA), parametric learning curve extrapolation (LCE), and the recently introduced LC-PFN model \cite{adriaensen2023efficient} for learning curve extrapolation.
We study these techniques in various classification and regression tasks for the class of fully connected deep neural networks.

Against all expectations, our findings are:
\begin{enumerate}
    \item dynamically allocating resources as done by successive halving or learning curve extrapolation offers minimal (and oftentimes no) utility compared to a constant number of training epochs, and
    \item one can often early discard models after only one epoch without losing significant final predictive performance, indicating that perhaps learning curves are more well-behaved than one may expect. 
\end{enumerate}
We believe these findings highlight the necessity to incorporate $1$-Epoch in future studies since it achieves such good predictive performance for minimal computation while being extremely simple to implement. The software used for our experiments is made publicly available\footnote{Code: \href{https://github.com/fmohr/lcdb/blob/ce96fa3768da94d222644883a11403119844f241/publications/2024-neurocom/multi-fidelity-hpo/README.md}{\texttt{https://github.com/fmohr/lcdb/blob/ce96fa3768da94d222644883a11403} \texttt{119844f241/publications/2024-neurocom/multi-fidelity-hpo/README.md}}}.

\section{Related Work}~\label{sec:related-work}

Our study focuses on methods, that train only a single model at a time, but keep all checkpoints for further reference. 
Early discarding means switching to training a model with another HP configuration before attaining the maximum number of epochs allowed. 
Such strategies are sometimes referred to as ``vertical'' model selection~\cite{mohr_learning_2022}. 

One well-known example is Asynchronous Successive Halving~\cite{li_system_2020} (SHA). 
Hyperband~\cite{li2017hyperband} can also be adapted to this setting, which can explore different trade-offs for SHA hyperparameters. Note that since we try different hyperparameters of SHA, Hyperband cannot improve over SHA in terms of the Pareto front, because Hyberband must incur some overhead. After all, it runs multiple versions of SHA inside, which is why it is not included in this comparison. 

Learning Curve Extrapolation~\cite{domhan_speeding_2015} (LCE) observes early performances and extrapolates them to decide whether training should continue. Learning Curve with Support Vector Regression~\cite{baker_accelerating_2017} predicts the final performance based on the configuration and early observations. Learning Curve with Bayesian neural networks ~\cite{klein_learning_2017} instead uses a Bayesian neural network.  Trace Aware Knowledge-Gradient~\cite{wu2020practical}  leverages an observed curve to update the posterior distribution of a Gaussian process more efficiently. \cite{adriaensen2023efficient} {uses a prior-fitted network \cite{hollmann2022tabpfn}, which is a transformer, to extrapolate learning curves, which is a sped-up and improved version of \cite{domhan_speeding_2015}}.  \cite{ruhkopf2022masif} extrapolates learning curves using a transformer to larger fidelities to predict the best algorithm from a portfolio, but does not perform regression.
\cite{mohr2023fast} uses a purely linear extrapolation, which is a conservative technique that is guaranteed to not prune the optimal candidate given the convexity of the learning curve.
The latter, however, is usually not the case for learning curves of neural networks.

FABOLAS~\cite{klein2017fast} uses a similar technique, where correlations are learned in the candidates' ranking between different levels of fidelity. Bayesian Optimization Hyperband~\cite{falkner_bohb_2018} embeds Bayesian optimization in Hyperband to sample candidates more efficiently. 

Some previous works sometimes implicitly make strong assumptions about the learning curve. For example, methods based on SH or SHA (implicitly) assume that the discarded learning curves will not cross in the future, since only Top-$K$ models are allowed to continue at any given step. In this context, models that start slowly are often discarded. 
This phenomenon is known as the ``short-horizon bias''~\cite{wu2018understanding}, and this is one of the most pressing reasons to introduce more complex models to deal with learning curves and their possibility of crossing.   
That is essentially what LCE methods aim to achieve. They either assume a parametric model \cite{domhan_speeding_2015,adriaensen2023efficient}, Gaussian process model \cite{wu2020practical}, complex hierarchical Bayesian models \cite{domhan_speeding_2015}, or Bayesian neural network models \cite{klein_learning_2017} to model the learning curves, to name a few examples. These models can make quite strong assumptions about the learning curves. 

It is not clear how often learning curves cross in general \cite{viering2022shape,mohr2023lcdb} and what kind of problem this poses for HPO.
This work investigates how often curves cross: if curves often cross, 1-epoch cannot perform well, generally, because it would discard too many slow-starting models. Our method further avoids making any assumptions about the learning curves, in the same spirit as SHA. One can see 1-Epoch as SHA to the extreme: where the reduction factor is set in such a way as to prune all models in one go. 

Benchmarks play a critical role in the design and development of HPO methods. We have surveyed several recent benchmarks for continuously evolving learning curves, such as HPOBench~\cite{eggensperger_hpobench_2021,klein_tabular_2019}, LCBench~\cite{zimmer2021auto}, JAHS-Bench-201~\cite{bansal2022jahsbench}, and YAHPO-Gym~\cite{pfisterer2022yahpo}.
In a preliminary version of this study~\cite{egele2023one}, we have already provided visualization and early elimination experiments for these different benchmarks that are consistent with this study.
However, as LCBench only had learning curves of 50 epochs and performance estimates on a test set, JAHS-Bench-201 and YAHPO-Gym are using a surrogate model which makes learning curves smoother, we prefer to use actual learning curve data to improve reliability.
Therefore, we have chosen to only use learning curves from 4 regression tasks in HPOBench~\cite{eggensperger_hpobench_2021,klein_tabular_2019}, and we resort to generating our own learning curves for classification with an experimental setup close to HPOBench. 

\section{Methods}
\label{sec:method}

\begin{figure}[!h]
     \centering
     \includegraphics[width=0.8\textwidth]{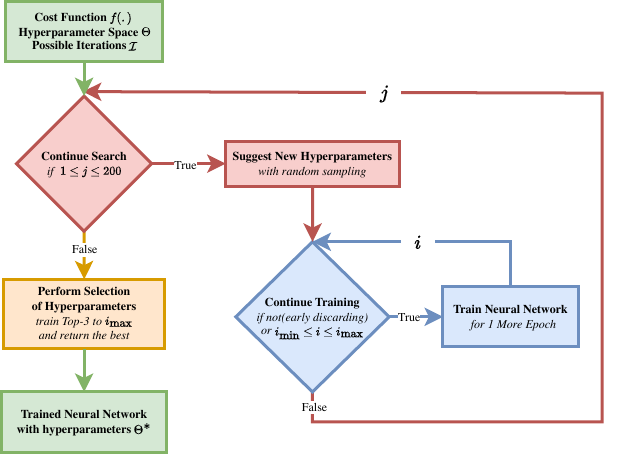}
     \caption{Hyperparameter optimization and its components including {\color{green}input/output}, {\color{red} outer optimization loop exploring new hyperparameter configurations}, {\color{blue}inner optimization loop incrementally allocating training iterations} ({\bf what we study in this work}) and {\color{orange} selection of hyperparameters to return}. In {\it italic} we specify the blocks to match with our experimental study.}
     \label{fig:components-hpo}
\end{figure}

We consider a function $f(\theta,i) \in \mathbb{R}$ that returns (empirical) generalization error of a deep neural network pipeline configured with hyperparameters $\theta \in \Theta$ (i.e., a vector of mixed variables) after $i \in \mathcal{I}$ training epochs. 
In our setting we bound the number of training epochs $i_\text{min} \leq i \leq i_\text{max}$.
Next consider a hyperparameter optimization algorithm $a \in \mathcal{A}$ such that $a(f,\Theta,\mathcal{I}) = (y_L, y_I)^T$ where $y_L = f(\theta^*,i_\text{max}) \in \mathbb{R}$ is the generalization error of the returned trained deep neural network pipeline configured with hyperparameters $\theta^*$ and $y_I \in \mathbb{N}$ is the total number of training epochs used by $a$ to complete the hyperparameter optimization process.
Then, the multi-objective problem that hyperparameter optimization with early discarding algorithms aims to solve is:

\begin{align}
    \label{eq:moo-optimization-problem}
    \min_{a \in \mathcal{A}} ~~ & (y_L, y_I) \\
             \textrm{s.t.  } ~~ & (y_L, y_I)^{T} = a(L,\Theta,\mathcal{I}) \nonumber
\end{align}

In Figure~\ref{fig:components-hpo} we provide a flowchart diagram of the hyperparameter optimization with early discarding algorithm class $\mathcal{A}$ that we consider.
The HPO process comprises an outer open cycle (red parts), in which an optimizer decides whether optimization should be continued or not.
If so, it picks a candidate hyperparameter (HP) configuration (or various if parallelization is supported) for evaluation.
Then, the performance of the chosen configurations is computed (blue parts).
Since we only consider training of neural networks one can think of the candidate evaluation as an inner cycle in which an empirical learning curve is constructed, with one entry per epoch.
In the orange box, a set of final candidates is selected (possibly of size 1) and trained to convergence (if not done already).
Among these, the candidate with the best performance is returned and serves as a trained model for predictions.

In the interest of separation of concerns, this paper focuses only on the aspect of early discarding (blue diamond).
The other components are fixed as follows:
The outer cycle simulates a random search with an evaluation limit of 200 pipelines, which are sampled offline to make sure that all early discarding methods decide upon the same pipelines.
Since no model of the performance landscape is built in the random search, the evaluation module simply returns the prediction performance of the network at the time when training is being stopped (no matter whether prematurely or because it has converged).
The orange component selects the 3 best configurations found during optimization and trains them to convergence (if not yet).
It then returns the best of these models.

This being said, our study focuses only on early discarding techniques for \emph{single candidates} as opposed to \emph{candidate portfolios}.
Many popular optimizers consider entire portfolios of candidates, which are then reduced at some predefined ratio \cite{jamieson2016non,li_hyperband_2018,falkner_bohb_2018,awad_dehb_2021}.
We are interested in a more flexible class of early discarding techniques that do not need to know all the candidates upfront but decide only upon one candidate at a time based on the score of the best candidate seen so far.
This is also referred to as the difference between horizontal optimization (simultaneously growing learning curves of a portfolio) and vertical optimization (evaluating candidates one by one, possibly without even knowing the whole set of candidates to be evaluated) \cite{mohr_learning_2022}.

Among these early discarding techniques for single candidates, we consider three state-of-the-art approaches from different research branches and an approach that simply trains the networks for a previously defined constant number of epochs.
First, for the idea of Successive Halving, which is used in many horizontal optimizers \cite{jamieson2016non,li_hyperband_2018,falkner_bohb_2018,awad_dehb_2021}, there is a sequential variant \cite{li_system_2020} that can be used as an independent early discarding module.
The second and third approaches discard candidates based on extrapolated learning curves using Monte Carlo Markov Chains (MCMC) \cite{egele2023one} and Prior Fitted Networks (PFN) \cite{adriaensen2023efficient}, respectively.
Another approach for extrapolation, learning curve-based cross-validation (LCCV) \cite{mohr_fast_2021} with state-of-the-art results in early discarding is not considered in the evaluation because it is based on the assumption of convexity (or concavity) of the learning curves, which is the typical case for sample-wise learning curves but not iteration-wise learning curves as created during the training of a neural network \cite{mohr_learning_2022}.

\subsection{Vertical Version of Successive Halving}
\label{sec:method-sha}
{\bf Successive Halving} (SHA)~\cite{jamieson2016non} is an \emph{optimization} technique that receives a \emph{set} of candidates, which is successively reduced while granting more resources to candidates that are being retained.
A common approach is to eliminate 50\% of the candidates and double the amount of resources for the remaining candidates until only one candidate remains; thereby, all iterations consume roughly the same quantity of compute resources.

It is possible to isolate the idea of SHA in order to use it as an early discarding module such as shown in blue in Figure~\ref{fig:components-hpo}~\cite{li_system_2020}. To do this, one can test at epoch $i$ if the currently observed score is among the top-$100/r\%$ already observed in the past for other candidates at the same epoch $i$, where $r$ is called the reduction factor (e.g., $r=2$ for a reduction of 50\%). If this is the case, then the training is continued otherwise it is stopped. This condition is not checked at every training epoch but follows a geometric schedule.

\subsection{Parametric Learning Curve Extrapolation with adapted MCMC variant}
\label{sec:method-lce}

{\bf Parametric Learning Curve Extrapolation} (LCE)~\cite{domhan_speeding_2015} uses a parametric model to predict the continuation of a learning curve. 
The parametric functions used for this task are mostly power laws originating from physics research \cite{mohr_learning_2022}.
It is also common to consider linear combinations of such functions \cite{domhan_speeding_2015}.

To enable the reconfigurability of greediness, we are interested in \emph{probabilistic extrapolations}.
That is, the extrapolation technique should output a \emph{distribution} over learning curves rather than just a single one (usually the likelihood maximizer).
These distributions can be obtained by sampling from the posterior distribution, usually using a Bayesian approach \cite{domhan_speeding_2015,klein_learning_2017}.

However, we found that the above techniques suffer from instabilities, which is why we here use a technique called RoBER (Robust Bayesian Early Rejection) \cite{egele2023one}.
Instead of considering a linear combination of several parametric models, we only consider one, that is MMF4 which was found to work well in general for extrapolation by \cite{mohr2023lcdb}.
In addition, we do not use a pure Bayesian approach but instead combine optimization with Bayesian inference. That is, first, we fit the parametric model using Levenberg-Marquadt, which minimizes the mean squared error on the observed anchors of the learning curve. Afterward, we use these fitting parameters $\hat{\theta}$ to derive a data-driven prior of the form $\theta \sim N(\hat{\theta},1)$.
We use a Gaussian likelihood on the observed learning curve anchors with an exponential prior with scale parameter 1. This completely defines the posterior, which is sampled using Markov-Chain-Monte-Carlo. This allows us to sample the distribution of extrapolated values at the largest anchor. We compute this distribution for each currently observed learning curve. If this distribution indicates for a learning curve candidate that we are with probability larger than $\rho$ worse than the current best-observed learning curve value, the candidate is eliminated. The larger $\rho$, the more conservative: for example if $\rho = 0.9$, a candidate is only discarded if the probability that it under-performs the currently best one at the horizon is greater than or equal to 90\%.

\subsection{Extrapolation via Prior Fitted Networks (PFN)}
Prior Fitted Networks (PFN) are transformer networks that are being trained on synthetic tasks sampled from a so-called prior distribution~\cite{muller2021transformers}.
% Here, tasks are described as labeled datasets with unlabeled test points.
For a new task, the PFN does not only output a single prediction for each test point but a \emph{distribution}.

Due to their general nature, PFNs can also be used to predict distributions over learning curves.
A recent approach that reports results comparable to or better than the MCMC approach of \cite{domhan_speeding_2015} was presented in \cite{adriaensen2023efficient}.
In this approach, synthetic learning curves are sampled from a prior distribution over linear combinations of model classes; a subset of those suggested in \cite{domhan_speeding_2015} is used.
The authors of this network offer a pre-trained implementation\footnote{LC-PFN code: \url{https://github.com/automl/lcpfn}}, which comes with an API that allows extrapolations of learning curves out of the box.
Our experiments are based on this implementation.

Regarding LCE, one can define a confidence level $\rho$ and discard candidates only if the probability that the limit performance is worse than the best currently known solution is at least $\rho$.

\subsection{i-Epoch: Constant Number of Epochs}
\label{sec:method-epoch}

The last and simplest method is one of a constant number of epochs.
In this case, the number of epochs is defined a priori and does not depend on any observations made during the evaluation of the candidate.
In our experiments, we consider all numbers of epochs between 1 and 100 as possible limits.

This method is different from the others in that it does not necessarily train \emph{any} model to convergence \emph{during} the evaluation.
In all the other approaches, at least one network, namely the one that is believed to be best, is trained until convergence.
On the contrary, in the case of a constant number of epochs, even the best network is not (necessarily) trained to convergence during evaluation but only in the final selection phase (orange box in Fig.~\ref{fig:components-hpo}).
Of course, if the number of epochs configured is high, it can \emph{implicitly} happen that the networks converge during evaluation.
In particular, no early \emph{stopping} (mind the difference to early discarding) is used to stop training once the curve has flattened out, so training can even take more epochs than what would be observed with a standard early stopping approach.

\section{Experimental Design}
\label{sec:exp_setup}

Our experiments were designed to answer the following research questions (RQs) for the hyperparameter optimization of deep neural networks:

\begin{enumerate}
    \item[\textbf{RQ1}:] What is the anytime performance of the HPO process (i.e., when stopped at any iteration of the red loop in Figure~\ref{fig:components-hpo}) when run with the different early discarding techniques for extreme configurations of discarding aggressiveness (i.e., when stopping training at the earliest and at the latest)?
    
    \item[\textbf{RQ2}:] For each early discarding technique, what is its Pareto-frontier in terms of (1) final predictive performance (of the selected and trained hyperparameter configuration) and (2) total training epochs consumed in the HPO process, obtained when testing different settings of the method?

    \item[\textbf{RQ3}:] What does each method contribute to the Pareto frontier resulting from all techniques? This aims to see if methods complement each other in terms of attainable trade-offs and which algorithm offers the most diverse trade-offs.

    \item[\textbf{RQ4}:] How does 1-Epoch compare to other methods and how can we understand its surprisingly good performance? 
\end{enumerate}

Preempting the detailed results, we already summarize at this point that the answers to these questions might be in contrast to the expectations in two ways:
\begin{enumerate}
    \item While it is clear that 1-Epoch is Pareto-optimal (since one cannot be faster), one would expect that $i$-Epoch tends to develop a sub-optimal Pareto frontier (compared to other early discarding methods) as $i$ grows.
This is because, since $i$-Epoch does not react to the previous performance observations, there is an increasing risk that (unpromising) candidates are trained unnecessarily long so that the number of total training epochs in the HPO increases without generating any benefit.
In other words, for pretty much any $i > i_\text{min}$ for some small $i_\text{min}$, e.g., 5 or 10, one would expect that there are configurations of the other early discarding methods that Pareto-dominate $i$-Epoch.
The surprising insight of our experiments is that {\bf the simple $i$-Epoch policy is rarely ever Pareto-dominated by any other method}.
    \item While one would generally expect the maximally aggressive strategy 1-Epoch to deliver significantly sub-optimal results in predictive performance $y_L$, we show that generally {\bf there is little and sometimes \emph{no} possible improvement in predictive performance $y_L$ over the $1$-Epoch baseline}.
    In several cases, {\bf 1-Epoch is not only Pareto-optimal but strictly optimal}.
\end{enumerate}

\subsection{Learning Curves Benchmarks}

To be able to generalize conclusions from this work, we answer the questions on several datasets, both regression and classification, which displayed noticeable differences in the observed learning curves. However, we limited our study to the class of fully connected deep neural networks, still including a variety of hyperparameters (e.g., preprocessing, residual connections, regularization).

All learning curves used to benchmark early discarding techniques were computed and stored \emph{prior} to the experimentation.
We now describe this generating process.
All evaluated deep neural networks are trained for 100 epochs, which fixes $i_\text{min}=1$ and $i_\text{max}=100$.
For \emph{regression} tasks, we used an external benchmark of pre-computed learning curves from HPOBench~\cite{eggensperger_hpobench_2021,klein_tabular_2019}. 
The deep neural networks from this benchmark are similar to ours but were generated from 9 hyperparameters listed in Table~\ref{tab:regression-hp-search-space} and 4 datasets were used.

\begin{table}[!h]
% \small
\centering
% \resizebox{\textwidth}{!}{%
\begin{tabular}{|c|c|}
\hline
~~~~~~~~\textbf{Hyperparameters}~~~~~~~~ & \textbf{Choices}                            \\ \hline
Initial LR               & $\{0.0005, 0.001, 0.005, 0.01, 0.05, 0.1\}$ \\ \hline
Batch Size               & $\{8, 16, 32, 64\}$                         \\ \hline
LR Schedule              & $\{\text{cosine}, \text{fix}\}$             \\ \hline
Activation/Layer 1       & $\{\text{relu}, \text{tanh}\}$              \\ \hline
Activation/Layer 2       & $\{\text{relu}, \text{tanh}\}$              \\ \hline
Layer 1 Size             & $\{16, 32, 64, 128, 256, 512\}$             \\ \hline
Layer 2 Size             & $\{16, 32, 64, 128, 256, 512\}$             \\ \hline
Dropout/Layer 1          & $\{0.0, 0.3, 0.6\}$                         \\ \hline
Dropout/Layer 2          & $\{0.0, 0.3, 0.6\}$                         \\ \hline
\end{tabular}%
% }
\caption{Hyperparameter search space for regression benchmarks defined in HPOBench~\cite{eggensperger_hpobench_2022,klein_tabular_2019}.}
\label{tab:regression-hp-search-space}
\end{table}

Datasets were split into 3 folds. 
The training split was used to optimize the neural network weights for a fixed hyperparameter configuration.
The validation split was used to optimize the hyperparameter configurations and serves as an estimate of generalization performance.
The test split was used as a final set of data to provide an unbiased report of our results.
The data split was 60\% for training, 20\% for validation, and 20\% for testing in the regression tasks, which was dictated by the setup of \cite{eggensperger_hpobench_2021}.
In the classification tasks, we chose the split to be 80\% for training, 10\% for validation, and 10\% for testing.

For classification tasks, we generated a set of 1,000 randomly sampled hyperparameter configurations from a search space of 17 hyperparameters listed in Table~\ref{tab:classification-hp-search-space}.
The learning curve generation for each classification task required about 1 hour of computation on 400 parallel NVIDIA A100 GPUs on the Polaris Supercomputer at the Argonne Leadership Computing Facility.

\begin{table}[!h]
\centering
\resizebox{\textwidth}{!}{%
\begin{tabular}{|c|c|}
\hline
\textbf{Hyperparameters} & \textbf{Choices}                            \\ \hline
Activation Function & \begin{tabular}[c]{@{}c@{}}$\{ \text{none}, \text{relu}, \text{sigmoid}, \text{softmax}, \text{softplus}, \text{softsign},$ \\ $\text{tanh}, \text{selu}, \text{elu}, \text{exponential}\}$\end{tabular} \\ \hline
Activity Regularizer & $\{ \text{none}, \text{L1}, \text{L2}, \text{L1L2}\}$ \\ \hline
Batch Normalization & $\{\text{True}, \text{False}\}$ \\ \hline
Batch Size & $[1, 512]$ (log-scale) \\ \hline
Bias Regularizer & $\{ \text{none}, \text{L1}, \text{L2}, \text{L1L2}\}$ \\ \hline
Dropout Rate & $[0.0, 0.9]$ \\ \hline
Kernel Initializer & \begin{tabular}[c]{@{}c@{}} $\{ \text{random-normal}, \text{random-uniform}, \text{truncated-normal},$ \\ $\text{zeros}, \text{ones}, \text{glorot-normal}, \text{glorot-uniform}, \text{he-normal},$ \\ $ \text{he-uniform}, \text{orthogonal}, \text{variance-scaling} \}$ \end{tabular} \\ \hline
Kernel Regularizer & $\{ \text{none}, \text{L1}, \text{L2}, \text{L1L2}\}$ \\ \hline
Learning Rate & $[10^{-5}, 10^{1}]$ (log-scale) \\ \hline
Number of Layers & $[1, 20]$ \\ \hline
Number of Units &  $[1, 200]$ (log-scale) \\ \hline
Optimizer & $\{\text{SGD}, \text{RMSprop}, \text{Adam}, \text{Adadelta}, \text{Adagrad}, \text{Adamax}, \text{Nadam}, \text{Ftrl}\}$ \\ \hline
Regularizer Factor & $[0.0, 1.0]$ \\ \hline
Shuffle Each Epoch & $\{\text{True}, \text{False}\}$ \\ \hline
Skip Connections & $\{\text{True}, \text{False}\}$ \\ \hline
Transform Categorical & $\{\text{onehot}, \text{ordinal}\}$ \\ \hline
Transform Real & $\{\text{minmax}, \text{std}, \text{none}\}$ \\ \hline
\end{tabular}%
}
\caption{Hyperparameter search space for classification benchmarks.}
\label{tab:classification-hp-search-space}
\end{table}

For all these configurations we compute the training, validation, and test learning curves by collecting confusion matrices on predictions.
Accounting for hyperparameter configurations that resulted in failures (e.g., ``\texttt{nan}'' loss with overflow or underflow) we end up with about 850 correct learning curves for each classification dataset.
The diversity of evaluated tasks is provided through the number of samples, features, classes or targets, and the type of features (real or categorical) in Table~\ref{tab:datasets}.

\begin{table}[!h]
\centering
\resizebox{\textwidth}{!}{%
\begin{tabular}{ccccccc}
\toprule
Dataset (OpenML-Id) & \#Features & \#Samples & \begin{tabular}[c]{@{}c@{}}\#Classes\\ or \#Targets\end{tabular} & \begin{tabular}[c]{@{}c@{}}Real\\ Features\end{tabular} & \begin{tabular}[c]{@{}c@{}}Categorical\\ Features\end{tabular} \\
\midrule
% Regression datasets
Slice Localization (42973) & 380 & 53,500 & 1 & True & False \\
Protein Structure (44963) & 9 & 45,730 & 1 & True & False \\
Naval Propulsion (44969) & 14 & 11,934 & 1 & True & False \\
Parkinson's Telemonitoring (4531) & 20 & 5,875 & 2 & True & True \\
\midrule
% Classification datasets
MNIST (554) & 784 & 70,000 & 10 & True & False \\
Australian Electricity Market (151) & 8 & 45,312 & 2 & True & True \\
Bank Marketing (1461) & 16 & 45,211 & 2 & True & True \\
Letter Recognition (6) & 16 & 20,000 & 26 & True & False \\
Letter Speech Recognition (300) & 617 & 7,797 & 26 & True & False \\
Robot Navigation (1497) & 24 & 5,456 & 4 & True & False \\
Chess End-Game (3) & 36 & 3,196 & 2 & False & True \\
Multiple Features (Karhunen) (14) & 76 & 2,000 & 10 & True & False \\
Multiple Features (Fourier) (16) & 64 & 2,000 & 10 & True & False \\
Steel Plates Faults (40982) & 27 & 1,941 & 7 & True & False \\
QSAR Biodegradation (1494) & 41 & 1,055 & 2 & True & False \\
German Credit (31) & 20 & 1,000 & 2 & True & True \\
Blood Transfusion (1464) & 4 & 748 & 2 & True & False \\
\bottomrule
\end{tabular}
}
\caption{Characteristics of datasets used for our experiments. On Top, the 4 datasets were used for regression, and on the bottom, the 6 datasets were used for classification. The datasets are sorted by decreasing number of samples.}
\label{tab:datasets}
\end{table}

\subsection{Experimental Protocol}

As we are interested in evaluating early discarding techniques (blue diamond in Figure~\ref{fig:components-hpo}) isolated from the process which suggests hyperparameter configurations, we propose the following experimental protocol.
The simulated process that suggests hyperparameter configurations (red rectangle in Figure~\ref{fig:components-hpo}) is a random sampling from the set of pre-computed learning curves.
This process is fixed by an initial random seed to simulate the same stream of candidate learning curves to different early discarding techniques.
The number of search iterations (red diamond in Figure~\ref{fig:components-hpo}) is fixed to 200 (main constant which makes outcomes of all experiments comparable).
Once the (red) loop of 200 candidates is over, the Top-3 models observed are selected and trained to completion if not already done.
A model that was not trained to completion during the previous 200 iterations will be retrained from scratch.
Of course, these additional training epochs are accounted for in the total number of training epochs used by the method.
For example, in 1-Epoch after 200 iterations we select the Top-3 candidates based on the observed scores $y_L$, we train them to completion so it consumes an additional $3 \times 100$ epochs, then we return the best from these 3. 
For 100-Epoch, as all evaluated models are already trained to completion no additional training is required.
The performance we report corresponds to the score reached by ``\emph{Method + Top-3}'' at any iteration of the search.
This corresponds to looking at the ``\emph{any-time}'' performance of each early discarding method, that is looking at what would be the outcome of the method if we were to stop after $k$ hyperparameter search iterations (red loop) for all $k\in[i_\text{min}=1, i_\text{max}=100]$.
A fixed set of 10 random seeds is set to perform 10 repetitions for each method.
This protocol ensured that each method was exposed to the same streams of candidates.
Therefore the different outcomes observed are only coming from differences in the decisions taken by each method to stop or continue the training.

\subsection{Performance Indicators}
\label{sec:performance-indicators}

In this section, we describe the two performance indicators of importance in our study. First, we detail the $R^2$ metric (generalized to both regression and classification) used to assess the predictive performance of evaluated hyperparameter configurations. Then, we detail the hypervolume indicator (HVI) metric used to assess the quality of the solutions for multi-objective optimization.

% Paragraph on R2 metric for regression and classification
First, we introduce the coefficient of determination $R^2$ in the case of regression tasks, where the target prediction is a real value, and, then we extend the notion to the case of classification tasks, where the target prediction is a categorical value in the spirit of \cite{el2017prediction}, also called the Prediction Advantage. 
This metric is useful as it standardizes both regression and classification similarly which helps us homogenize regression and classification learning curves.
A dataset $D = \{(x_1,y_1), ..., (x_n,y_n)\}$ is composed of i.i.d. variables from the joint distribution $P(X,Y)$. 
In \emph{regression}, the usual definition of $R^2$ (a.k.a., coefficient of determination) is:
\begin{equation}
\label{eq:classic-regression-r2}
R^2 := 1 - \frac{SS_{res}}{SS_{tot}} = 1 - \frac{\sum_{i=1}^n (y_i-\hat{y}(x_i))^2}{\sum_{i=1}^n (y_i-\bar{y})^2}
\end{equation}
where $SS_{res}$ is the residual sum of squares, $SS_{tot}$ is the total sum of squares, $\bar{y} = \frac{1}{n}\sum_{i=1}^n y_i$ is the empirical mean of the marginal distribution $P(Y)$ and, $\hat{y}(x_i)$ is a prediction from our model.
This definition can also written as:
\begin{equation}
\label{eq:regression-r2-with-variance}
R^2 = 1 - \frac{\frac{1}{n}\sum_{i=1}^n L_\text{2}(y_i,\hat{y}(x_i))}{\frac{1}{n}\sum_{i=1}^n  L_\text{2}(y_i,\bar{y})} \approx 1 - \frac{E[(Y-E[Y|X])^2|X]}{E[(Y-E[Y])^2]} 
\end{equation}
In the form given by Equation~\ref{eq:regression-r2-with-variance}, the expectations $E[Y]$ and $E[Y|X]$ correspond to the optimal Bayes predictors for the squared loss $L_\text{2}(Y,\hat{Y}) = (Y-\hat{Y})^2$ respectively on the marginal and conditional distributions. 
Therefore $R^2$ corresponds to the normalization of the expected error of the optimal Bayes predictor on the conditional $P(Y|X)$ distribution by the expected error of the optimal Bayes predictor on the marginal distribution $P(Y)$ (a.k.a., constant or ``dummy'' predictor).
In \emph{classification}, we replace the squared-loss with the $0-1$ loss $L_\text{0-1}(Y,\hat{Y}) = 1 \text{ if } Y \neq \hat{Y} \text{ else } 0$. The optimal Bayes predictor becomes the mode instead of the mean (i.e., the class with the highest probability). We then obtain a new definition of $R^2$ for classification:
\begin{equation}
R^2 = 1 - \frac{\frac{1}{n}\sum_i L_\text{0-1}(y_i,\hat{y}(x_i))}{\frac{1}{n} \sum_i L_\text{0-1}(y_i,\dot{y})}
\end{equation}
where $\dot{y}$ is the mode on the marginal distribution $P(Y)$. This is also known as the Prediction Advantage \cite{el2017prediction}. 
For both regression and classification, we have that performance of zero means that the model is as bad as the optimal constant predictor that only uses information from the marginal $P(Y)$ and ignores the input $X$. If the $R^2$ is 1 the prediction is ``\emph{perfect}'' (which also means that there is no presence of random noise on the target).
In our study, {\bf the goal is to maximize the $R^2$ score for improved predictive performance} which is equivalent to minimizing $y_L := 1 - R^2(\theta,i_\text{max})$ in Equation~\ref{eq:moo-optimization-problem} (replacing the $\mathcal{L}$ by our $R^2$ score).

% Paragraph on Hypervolume Indicator
Now that we have discussed the performance indicator for prediction we will present the metric used to assess the quality of multi-objective optimization (MOO).
For the sake of brevity, we will not recall the formal definitions related to the notion of Pareto-optimality in MOO.
However, shortly we recall that Pareto-Front refers to the solution set in the objective space (i.e., 2-dimensional in our case as we have 2 objectives $y_L$ and $y_I$).
As these objectives are (supposedly) conflicting, $y_L$ the predictive performance and $y_I$ the total number of training iterations used, the Pareto-Front is a one-dimensional space (i.e., a line) unless the problem is ``degenerated'', meaning there is no real conflict between objectives and the solution set is therefore containing a single point.
Among the possible metrics used in MOO~\cite{audet2021performance} and as we do not know the true Pareto-Front of our problem we decide to use the hypervolume indicator (HVI). 
As we are in 2-D it corresponds to measuring the area defined by an estimated Pareto-Front and a reference point (fixed for all experiments on the same dataset).
The HVI is compliant with the notion of Pareto-optimality and also known to measure the compare the diversity of solutions (i.e., trade-offs) between different Pareto-Fronts.
In our study, {\bf the goal is to identify the early discarding technique which maximizes the Hypervolume indicator when evaluated at different levels of aggressiveness}.

\section{Results}
\label{sec:results}

In this section, we present the results that helped us answer the research questions introduced in Section~\ref{sec:exp_setup}.

\subsection{RQ1 -- What is the anytime performance of early discarding techniques?}

\begin{figure}[!h]
    \centering
    \begin{subfigure}[b]{0.45\textwidth}
        \centering
        \includegraphics[width=\textwidth]{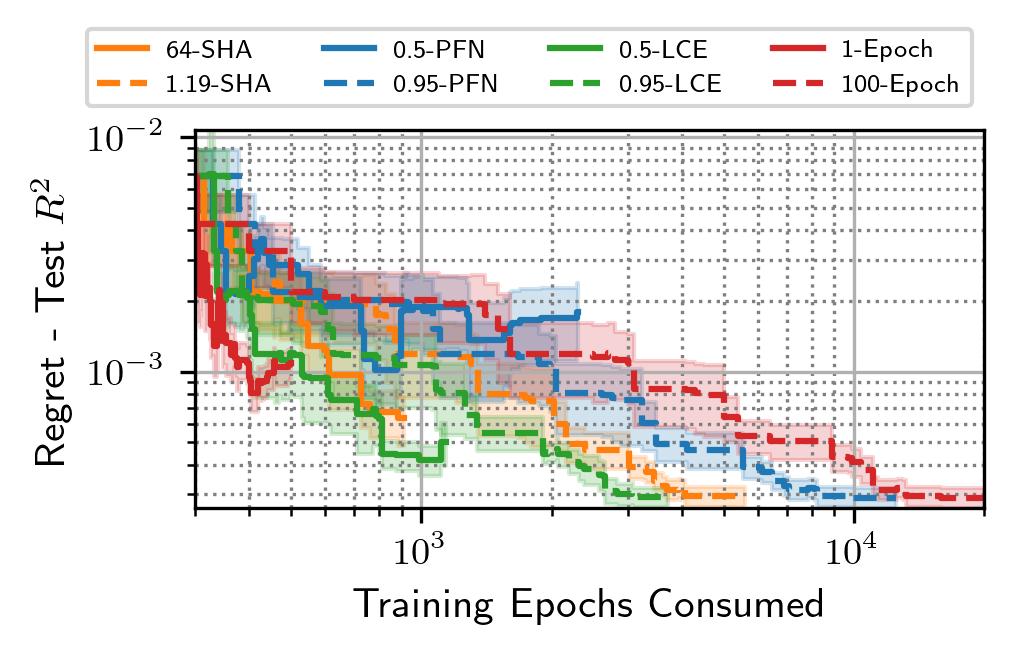}
        \caption{Slice Localization}
        \label{fig:curves-mfhpo-slicelocalization}
    \end{subfigure}
    \begin{subfigure}[b]{0.45\textwidth}
        \centering
        \includegraphics[width=\textwidth]{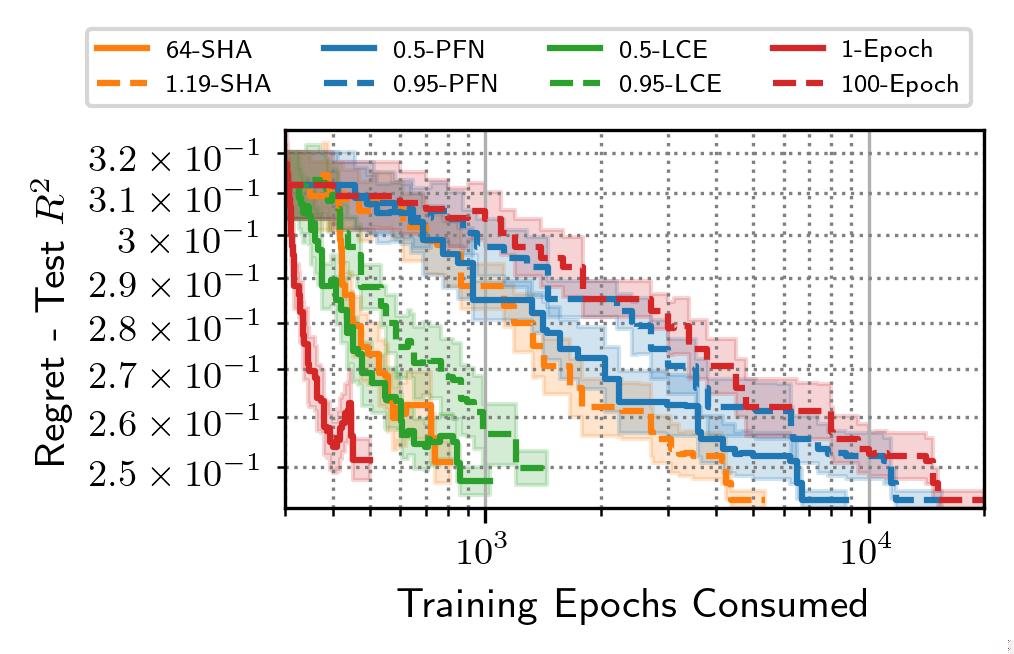}
        \caption{Protein Structure}
        \label{fig:curves-mfhpo-protein}
    \end{subfigure}\\
    \begin{subfigure}[b]{0.45\textwidth}
        \centering
        \includegraphics[width=\textwidth]{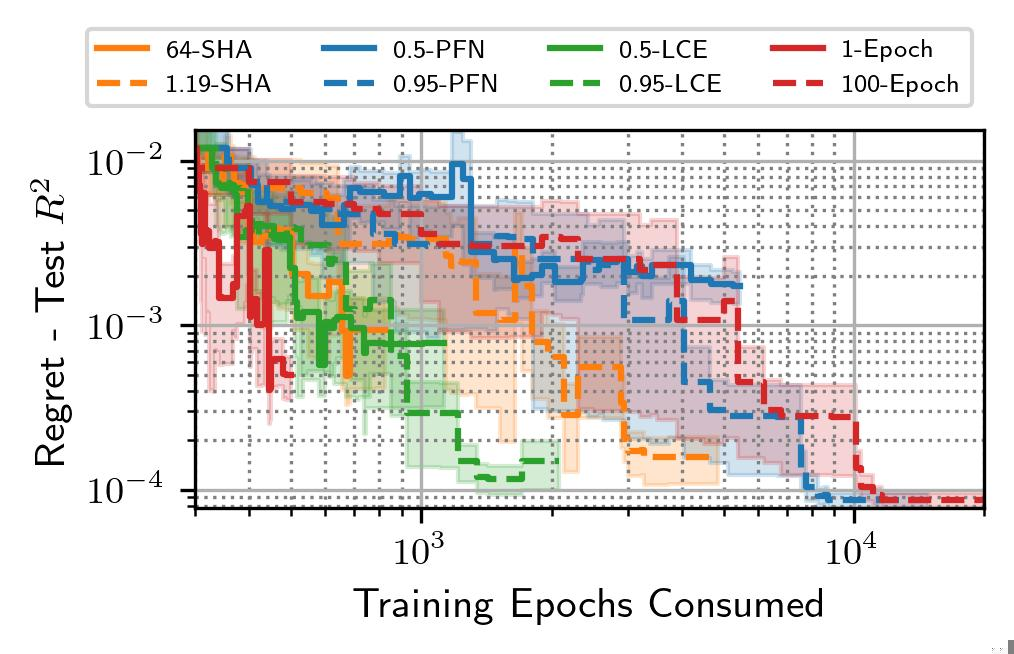}
        \caption{Naval Propulsion}
        \label{fig:curves-mfhpo-navalpropulsion}
    \end{subfigure}
    \begin{subfigure}[b]{0.45\textwidth}
        \centering
        \includegraphics[width=\textwidth]{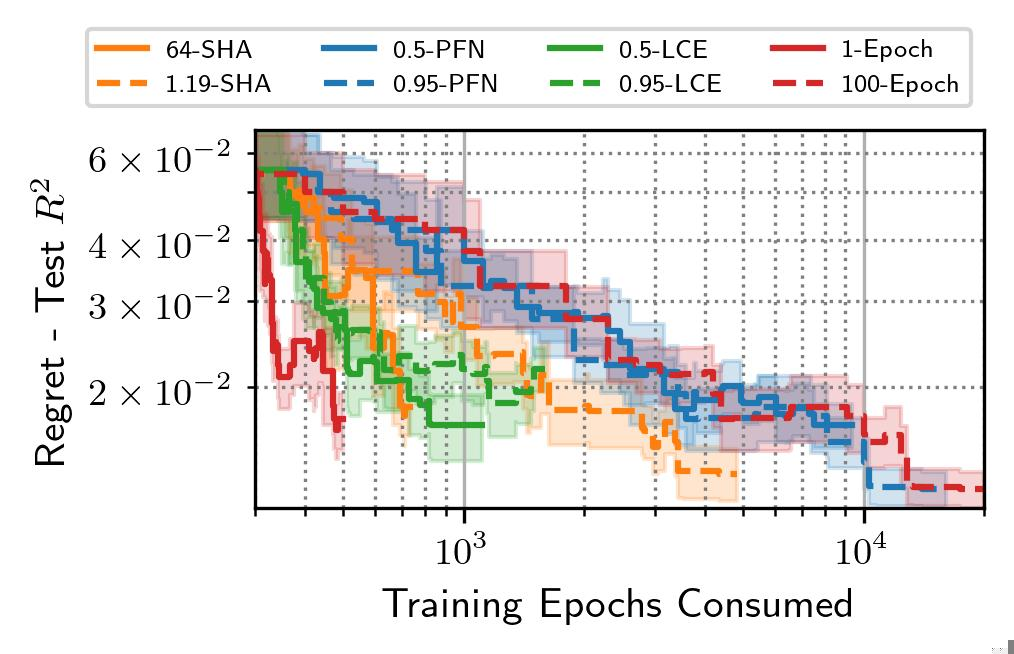}
        \caption{Parkinson's Telemonitoring}
        \label{fig:curves-mfhpo-parkinsons}
    \end{subfigure}
    \caption{Comparing the any-time performance of various early discarding techniques during a random search (mean and one standard error over 10 repetitions) of 200 iterations (4 regression tasks). The two baseline strategies $1$-Epoch and $100$-Epoch method bound the trade-offs that can be achieved. {\bf The predictive performance of $1$-Epoch is at least of the same order of magnitude as other strategies while consuming a significantly smaller (the minimum in training epochs) number of training epochs}.}
    \label{fig:curves-mfhpo-regression}
\end{figure}

\begin{figure}[!h]
    \centering
    \begin{subfigure}[b]{0.32\textwidth}
        \centering
        \includegraphics[width=\textwidth]{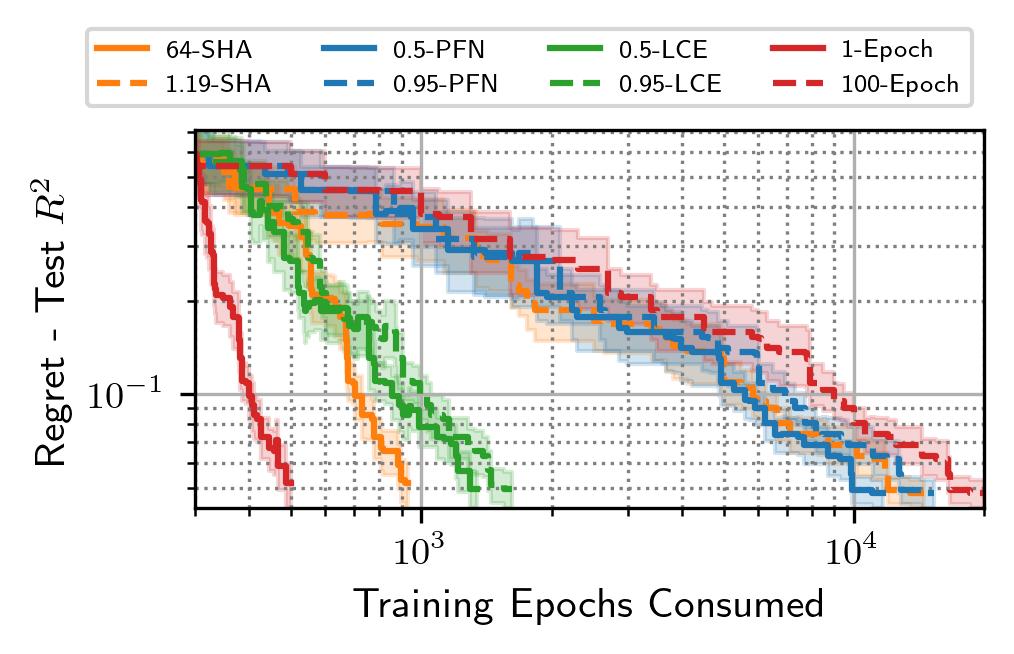}
        \caption{MNIST}
        \label{fig:curves-mfhpo-mnist}
    \end{subfigure}
    \begin{subfigure}[b]{0.32\textwidth}
        \centering
        \includegraphics[width=\textwidth]{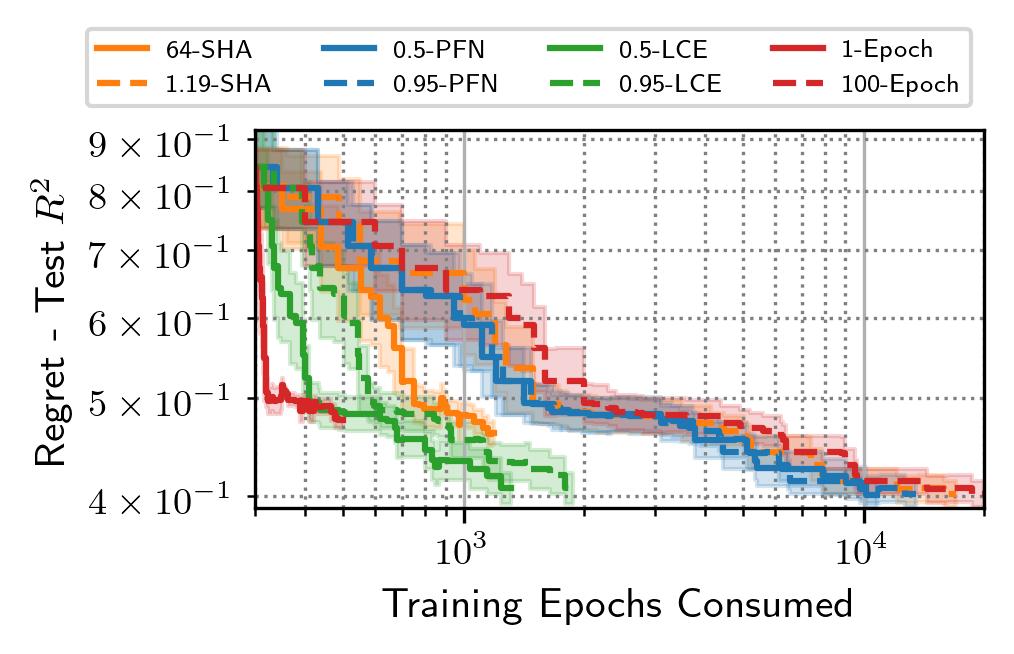}
        \caption{Australian Electricity Market}
        \label{fig:curves-mfhpo-electricitymarket}
    \end{subfigure}
    \begin{subfigure}[b]{0.32\textwidth}
        \centering
        \includegraphics[width=\textwidth]{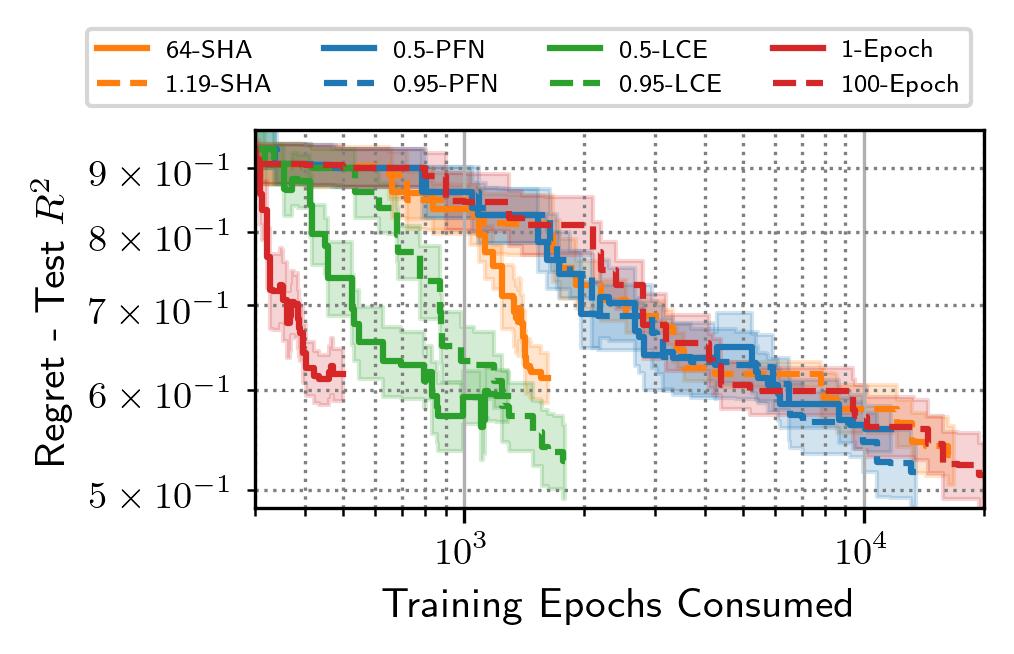}
        \caption{Bank Marketing}
        \label{fig:curves-mfhpo-bankmarketing}
    \end{subfigure}
    \begin{subfigure}[b]{0.32\textwidth}
        \centering
        \includegraphics[width=\textwidth]{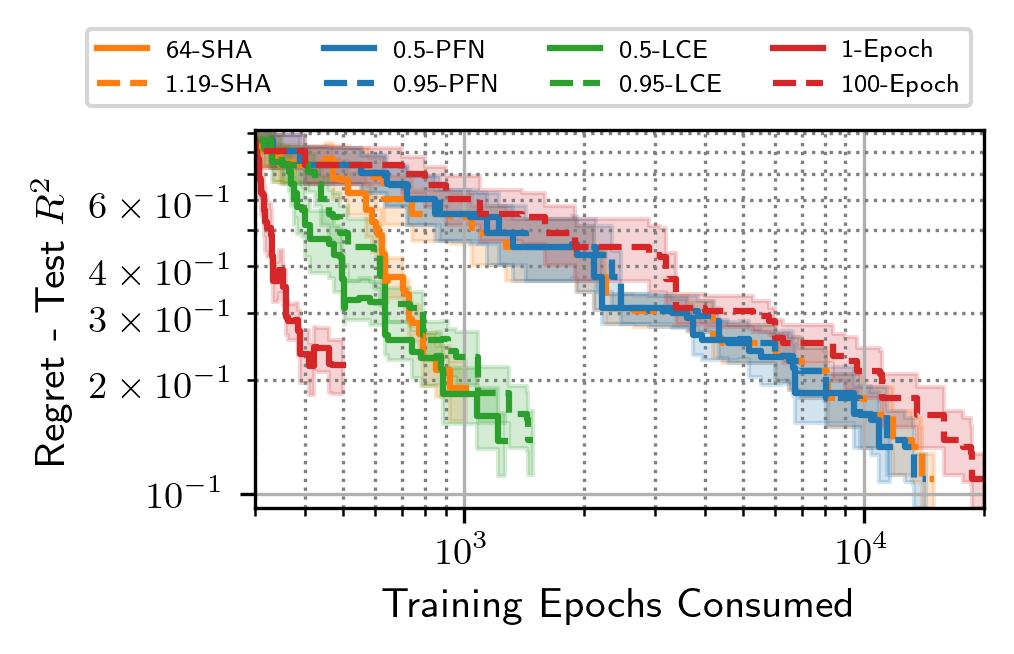}
        \caption{Letter Recognition}
        \label{fig:curves-mfhpo-letterrecognition}
    \end{subfigure}
    \begin{subfigure}[b]{0.32\textwidth}
        \centering
        \includegraphics[width=\textwidth]{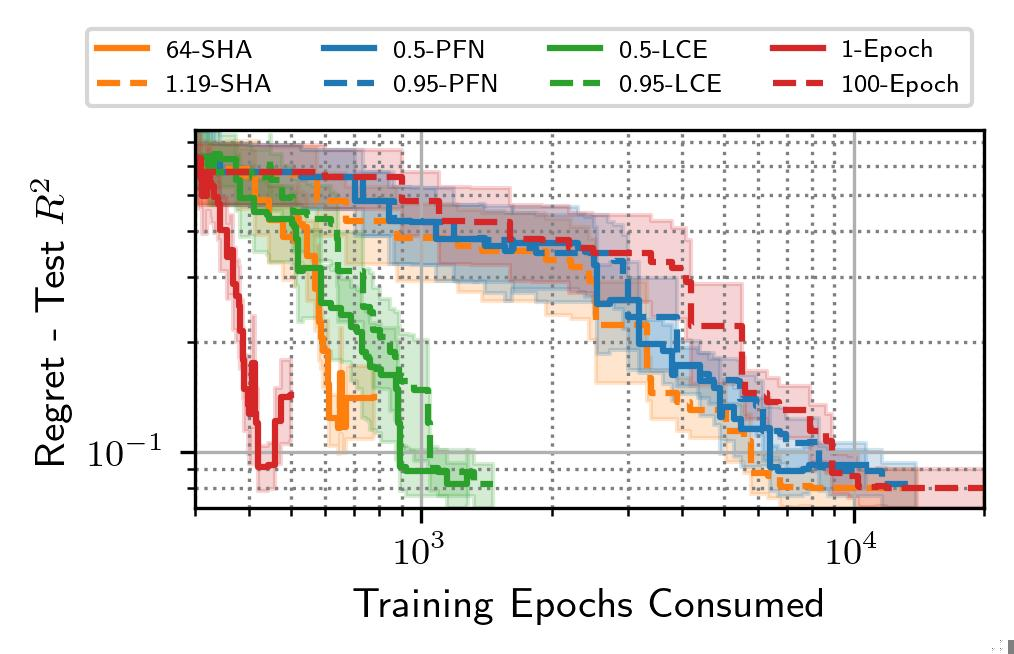}
        \caption{Speech Recognition}
        \label{fig:curves-mfhpo-speechrecognition}
    \end{subfigure}
    \begin{subfigure}[b]{0.32\textwidth}
        \centering
        \includegraphics[width=\textwidth]{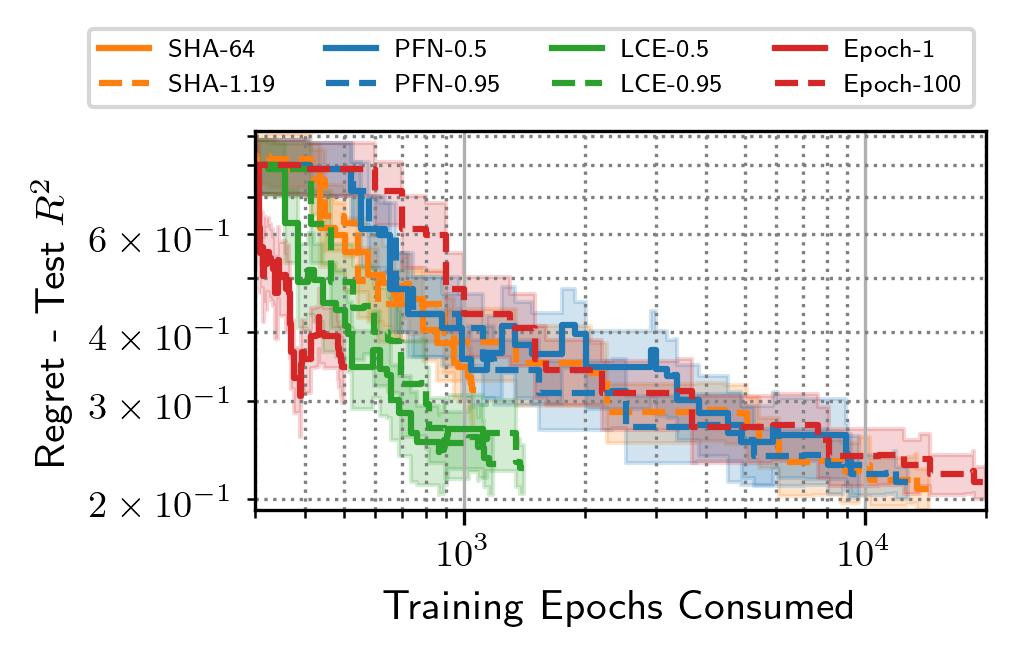}
        \caption{Robot Navigation}
        \label{fig:curves-mfhpo-robotnavigation}
    \end{subfigure}
    \begin{subfigure}[b]{0.32\textwidth}
        \centering
        \includegraphics[width=\textwidth]{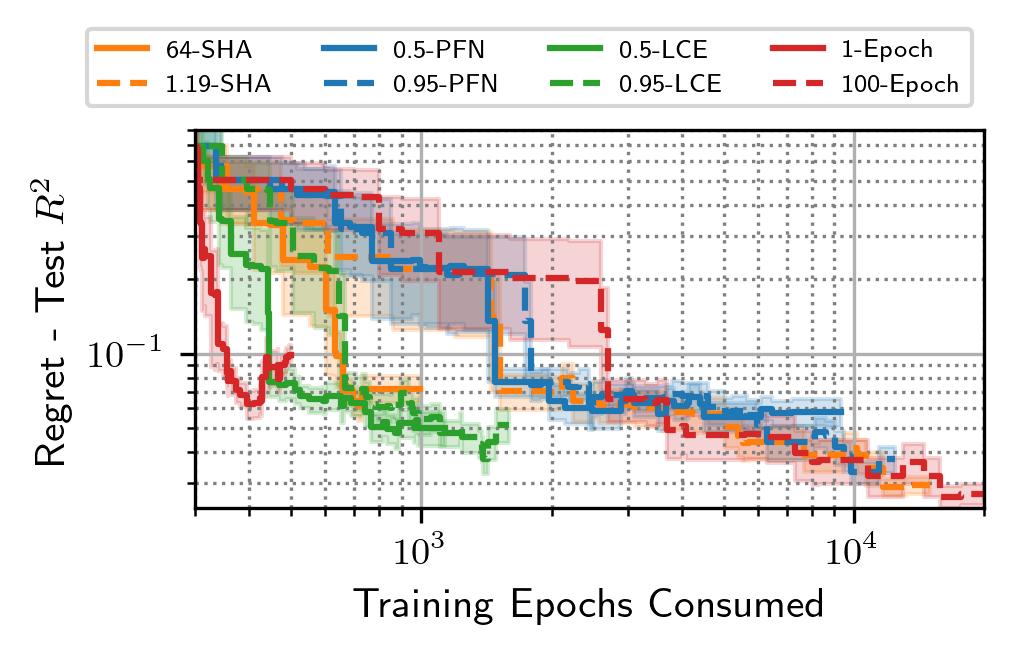}
        \caption{Chess End-Game}
        \label{fig:curves-mfhpo-chessendgame}
    \end{subfigure}
    \begin{subfigure}[b]{0.32\textwidth}
        \centering
        \includegraphics[width=\textwidth]{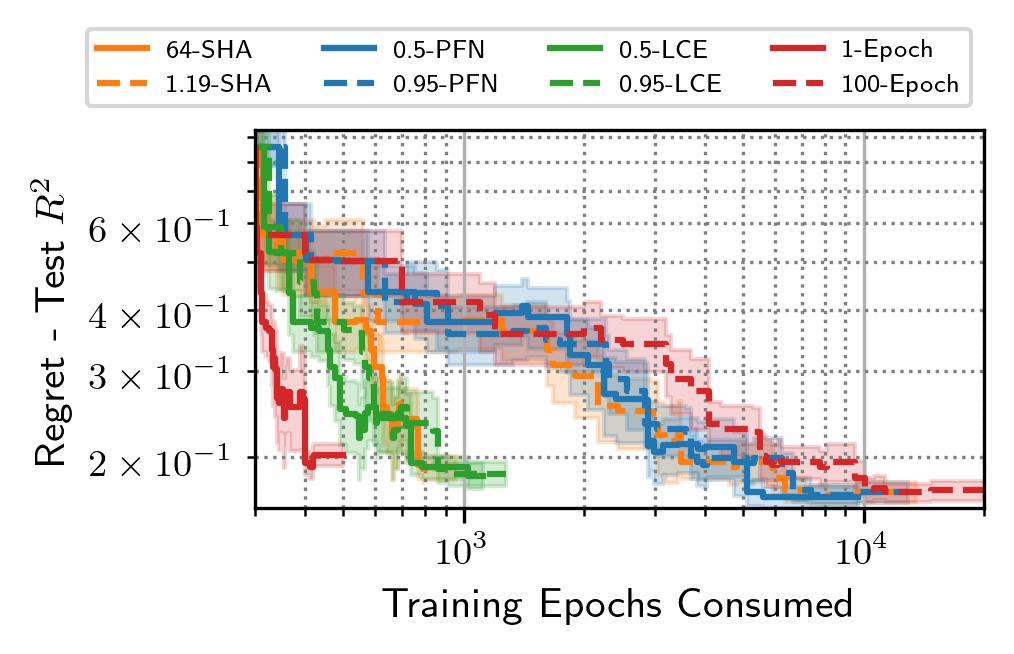}
        \caption{Multiple Features (Karhunen)}
        \label{fig:curves-mfhpo-mfkarhunen}
    \end{subfigure}
    \begin{subfigure}[b]{0.32\textwidth}
        \centering
        \includegraphics[width=\textwidth]{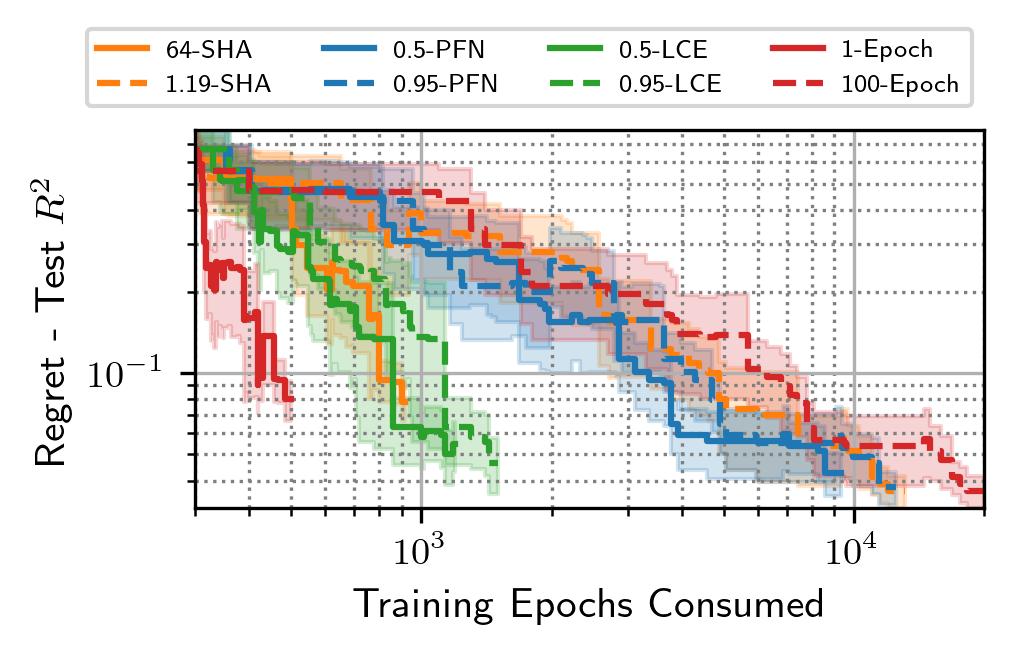}
        \caption{Multiple Features (Fourier)}
        \label{fig:curves-mfhpo-mffourier}
    \end{subfigure}
    \begin{subfigure}[b]{0.32\textwidth}
        \centering
        \includegraphics[width=\textwidth]{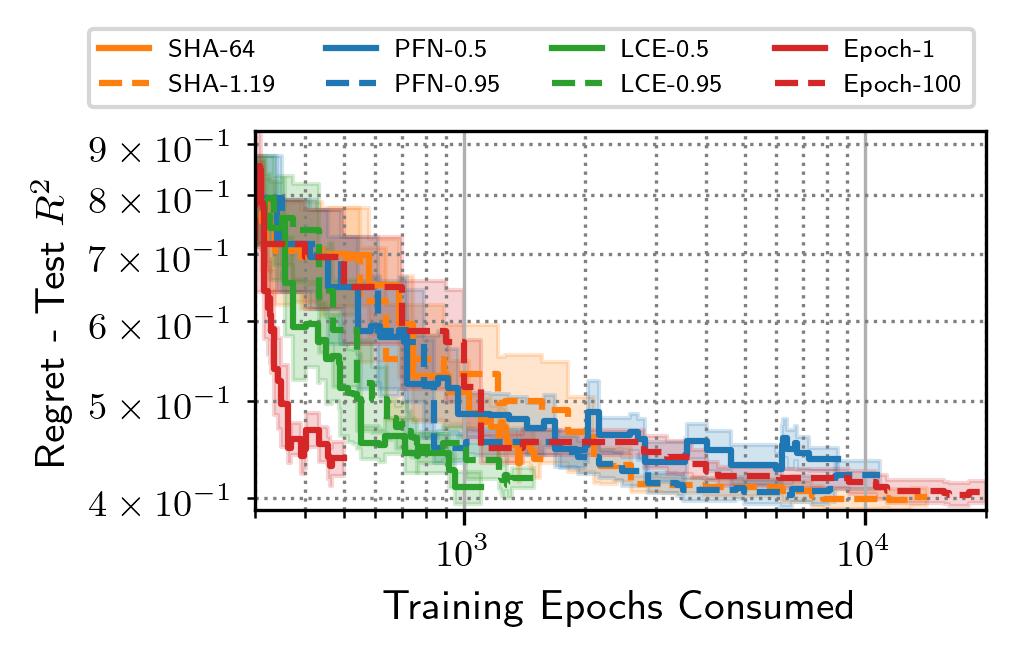}
        \caption{QSAR Biodegradation}
        \label{fig:curves-mfhpo-qsarbiodegradation}
    \end{subfigure}
    \begin{subfigure}[b]{0.32\textwidth}
        \centering
        \includegraphics[width=\textwidth]{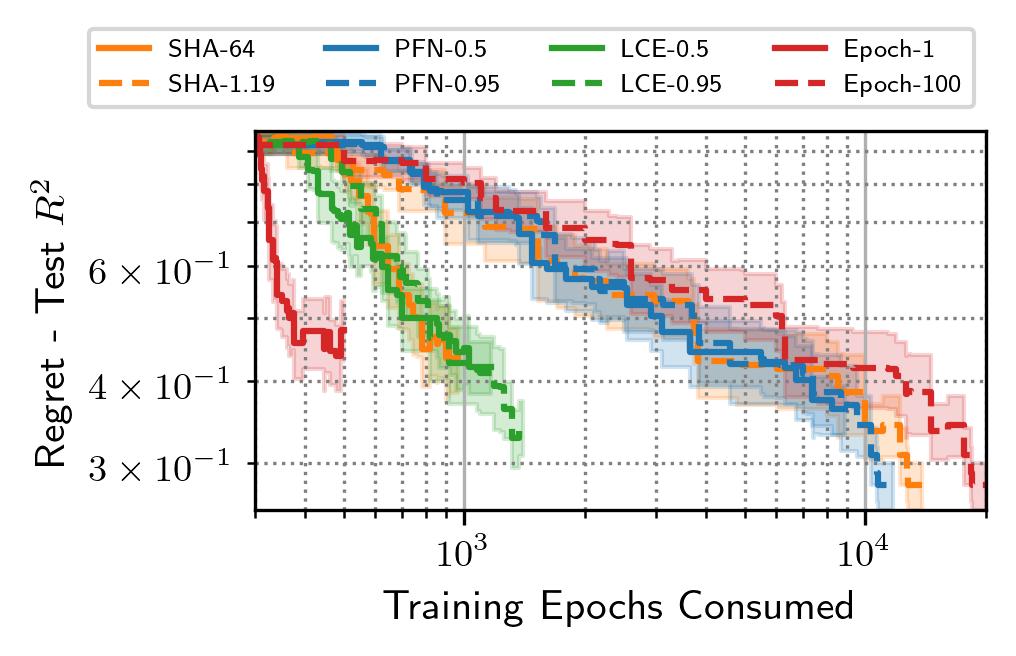}
        \caption{Steel Plates Faults}
        \label{fig:curves-mfhpo-steelplatesfaults}
    \end{subfigure}
    \begin{subfigure}[b]{0.32\textwidth}
        \centering
        \includegraphics[width=\textwidth]{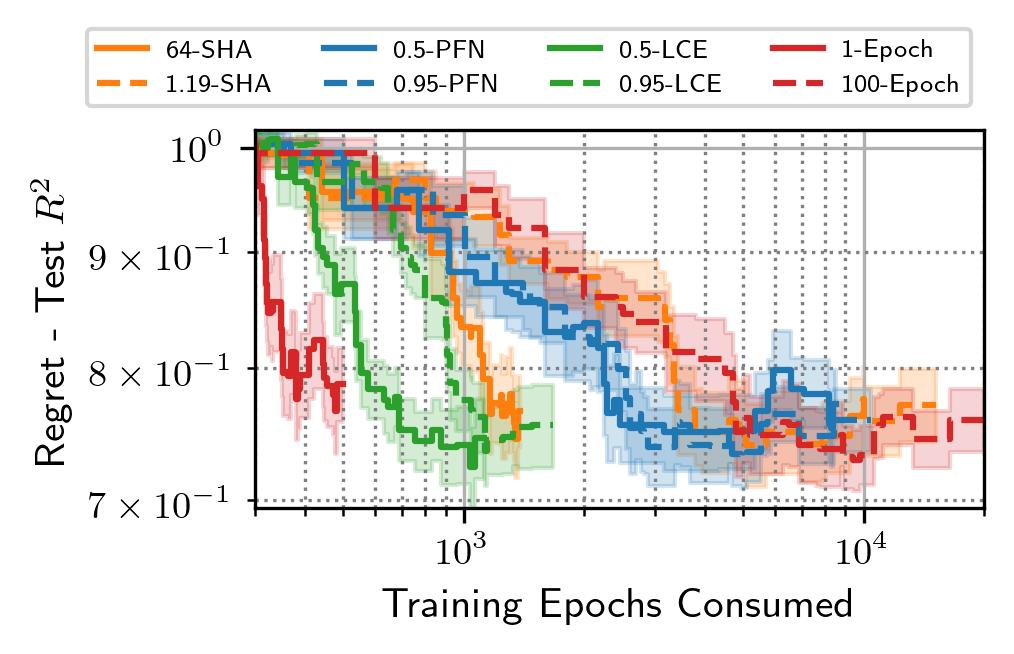}
        \caption{German Credit}
        \label{fig:curves-mfhpo-germancredit}
    \end{subfigure}
    \begin{subfigure}[b]{0.32\textwidth}
        \centering
        \includegraphics[width=\textwidth]{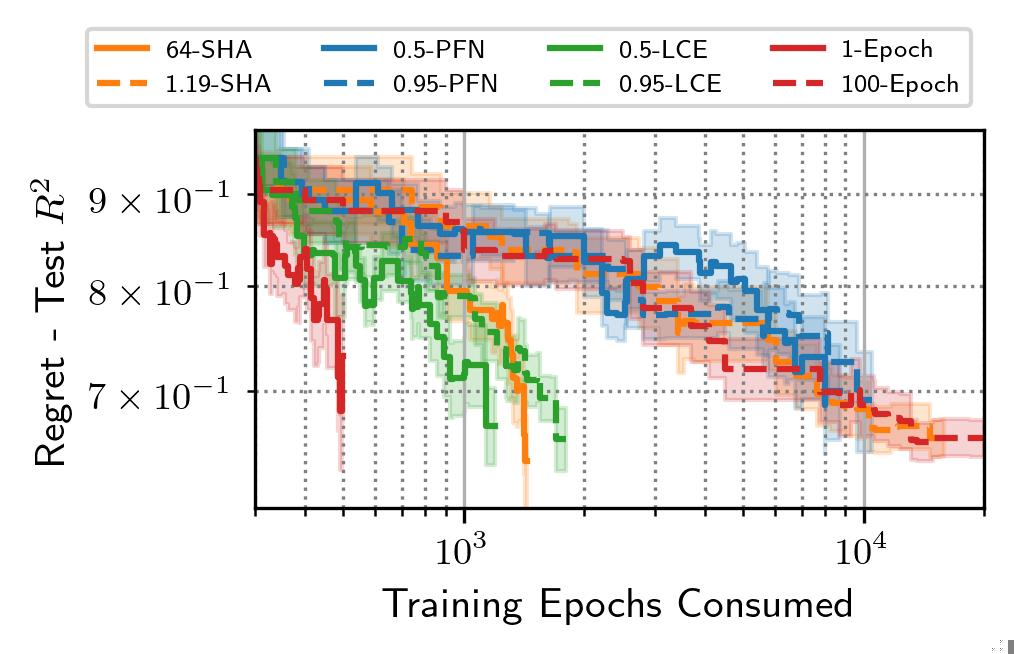}
        \caption{Blood Transfusion}
        \label{fig:curves-mfhpo-bloodtransfusion}
    \end{subfigure}
    \caption{Comparing the any-time performance of various early discarding techniques during a random search (mean and one standard error over 10 repetitions) of 200 iterations (on 13 classification tasks).}
    \label{fig:curves-mfhpo-classification}
    \vspace{-1em}
\end{figure}

To understand the anytime performance of early discarding techniques we plot the $1-R^2$ \emph{test} performance as a function of overall training epochs realized so far.
That is, a curve that passes the point $(t,l)$ in the plot means that the test score of the model that \emph{would} have been picked if the HPO process had stopped after $t$ total training epochs would have been $l$.
This type of performance curve weighs training epochs equally for all hyperparameter configurations, which may be deceiving since they can vary in computational cost (e.g., large and small neural networks).
Still, it is a convenient simple method abstracting from implementation details.
We present the performance curves in Figures~\ref{fig:curves-mfhpo-regression} and \ref{fig:curves-mfhpo-classification} for classification and regression respectively.

The most important insight from the plots is that the sensitivity of the early discarding techniques with respect to their aggressiveness parameter varies a lot. 
While the $i$-Epoch and $r$-SHA algorithms are very sensitive to the aggressiveness (as expected), the learning curve extrapolation-based methods (i.e., $\rho$-LCE and $\rho$-PFN) are surprisingly less sensitive to aggressiveness parameter $\rho$.
This can be observed especially on the set of classification tasks shown in Figure~\ref{fig:curves-mfhpo-classification}.
In other words, for $\rho$-LCE and $\rho$-PFN, it almost makes no difference in consumed training epochs whether the user requires almost certainty ($\rho = 0.95$) or whether the certainty is just as good as a coin flip ($\rho = 0.5$).
This could indicate that the models express too little uncertainty about the extrapolated learning curve.

% PFN does not reduce training iterations
Another observation is that the $\rho$-PFN method hardly reduces the overall training epochs used by 100-Epoch as can be seen for all datasets. It means that the learning curve extrapolation of this method is probably over-optimistic.
It even seems to perform worse than 100-Epoch for both predictive performance and overall training epochs used on learning curves which are very noisy and increasing. These failures can be observed in Figures~\ref{fig:curves-mfhpo-bankmarketing}, \ref{fig:curves-mfhpo-germancredit} and \ref{fig:curves-mfhpo-bloodtransfusion}.

% LCE can be under-performing in predictive accuracy even when the setting is conservative (0.95)
A third observation is that $\rho$-LCE, while being a more robust version of LCE, can still under-perform predictive performance even when being set to be conservative ($\rho$=0.95). This can be seen in Figures~\ref{fig:curves-mfhpo-parkinsons}, \ref{fig:curves-mfhpo-letterrecognition}, \ref{fig:curves-mfhpo-mfkarhunen} and \ref{fig:curves-mfhpo-steelplatesfaults}.
This confirms our belief that such models express too little uncertainty about the extrapolation.

% What would be the perfect solution in practice
From the practical viewpoint, no utopia method has yet been found.
A utopia method would achieve a strict and consistent dominance compared to the 100-Epoch baseline.
That is a method that achieves, on all tasks, better predictive performance while being faster than the base full training evaluation.
Such a method seems not to exist currently and may not exist if both objectives $y_L$ and $y_I$ are truly conflicting.

% The importance of including 1-Epoch to include the worst case scenario in accuracy
Finally, the presented performance curve plots also show the importance of considering the 1-Epoch baseline to contextualize results and avoid an overly optimistic presentation of the methods.
Without the solid red line which corresponds to 1-Epoch, $\rho$-LCE might appear a quite dominant approach in this experimental setting.
While it is true that learning curve extrapolation-based methods are very convincing in many cases, there are some datasets, such as Protein Structure (Figure~\ref{fig:curves-mfhpo-protein}), Parkinson’s Telemonitoring (Figure~\ref{fig:curves-mfhpo-parkinsons}), MNIST (Figure~\ref{fig:curves-mfhpo-mnist}), QSAR-Biodegradation (Figure~\ref{fig:curves-mfhpo-qsarbiodegradation}), or German Credit (Figure~\ref{fig:curves-mfhpo-germancredit}), in which the 1-Epoch baseline can reduce the number of epochs of LCE again by about 50\% without losing significant or any predictive performance.

\subsection{RQ2 -- Multi-objective trade-offs and Pareto-Fronts}

\begin{figure}[t]
    \centering
    \begin{subfigure}[b]{0.45\textwidth}
        \centering
        \includegraphics[width=\textwidth]{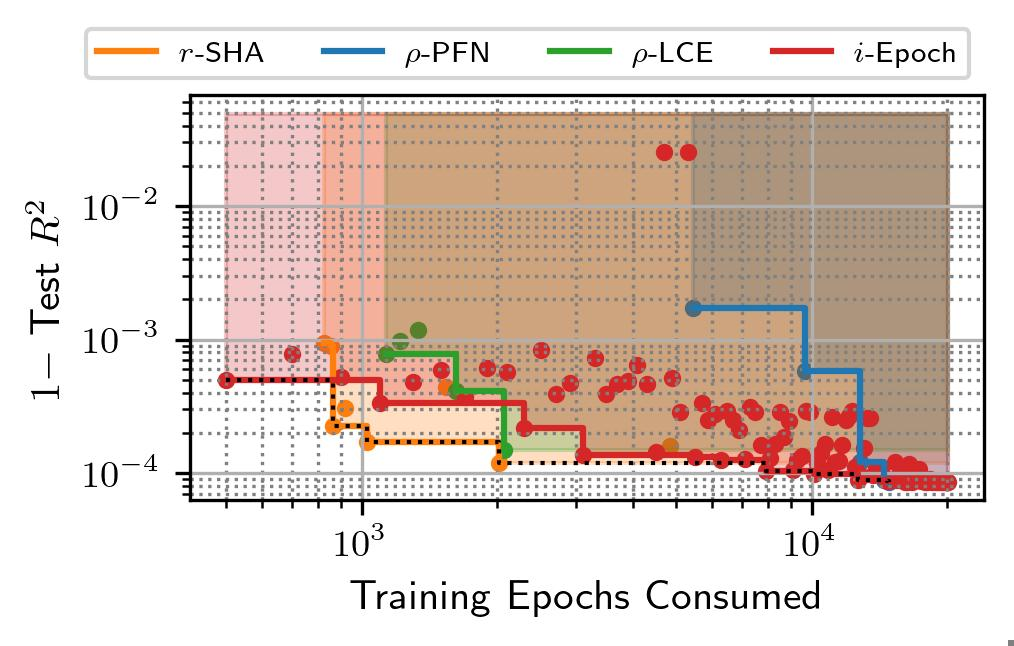}
        \caption{Naval Propulsion}
        \label{fig:paretofront-mfhpo-navalpropulsion}
    \end{subfigure}
    \begin{subfigure}[b]{0.45\textwidth}
        \centering
        \includegraphics[width=\textwidth]{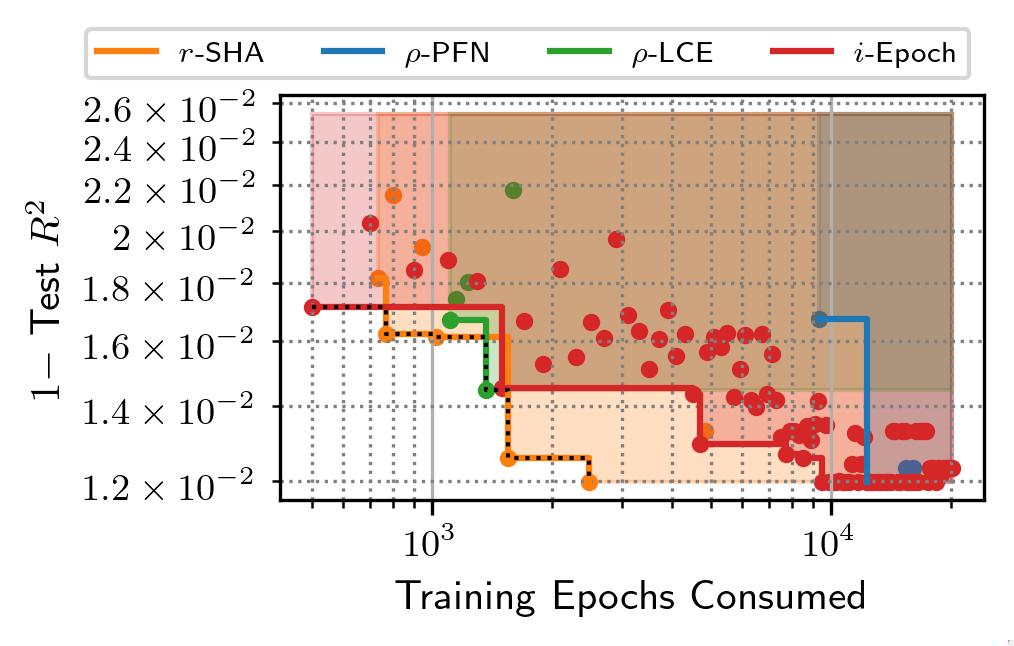}
        \caption{Parkinson's Telemonitoring}
        \label{fig:paretofront-mfhpo-parkinsons}
    \end{subfigure}\\
    \begin{subfigure}[b]{0.45\textwidth}
        \centering
        \includegraphics[width=\textwidth]{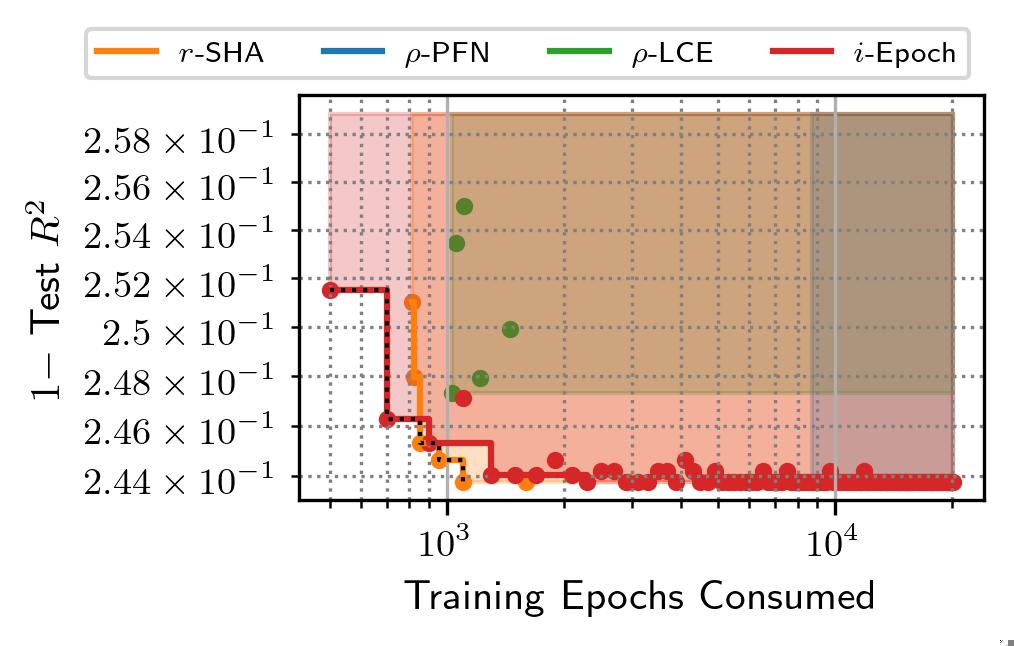}
        \caption{Protein Structure}
        \label{fig:paretofront-mfhpo-protein}
    \end{subfigure}
    \begin{subfigure}[b]{0.45\textwidth}
        \centering
        \includegraphics[width=\textwidth]{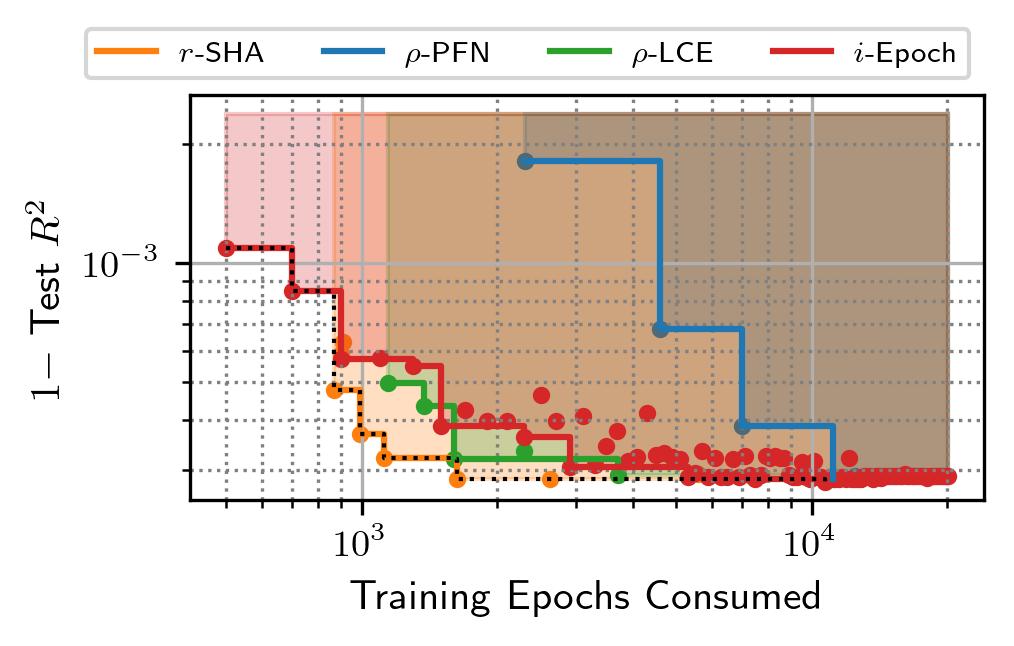}
        \caption{Slice Localization}
        \label{fig:paretofront-mfhpo-slicelocalization}
    \end{subfigure}
    \caption{Multi-objective profiles built from spanning various levels of aggressiveness of early discarding methods (on 4 regression tasks). The estimated Pareto-Front of each method is shown in a plane line. The black dotted line corresponds to the estimated Pareto-Front including the methods altogether. {\bf It can be seen that the $i$-Epoch strategy spans more trade-offs (larger area) than other methods while never being significantly dominated}.}
    \label{fig:paretofront-mfhpo-regression}
\end{figure}
\begin{figure}[!h]
    \centering
    \begin{subfigure}[b]{0.32\textwidth}
        \centering
        \includegraphics[width=\textwidth]{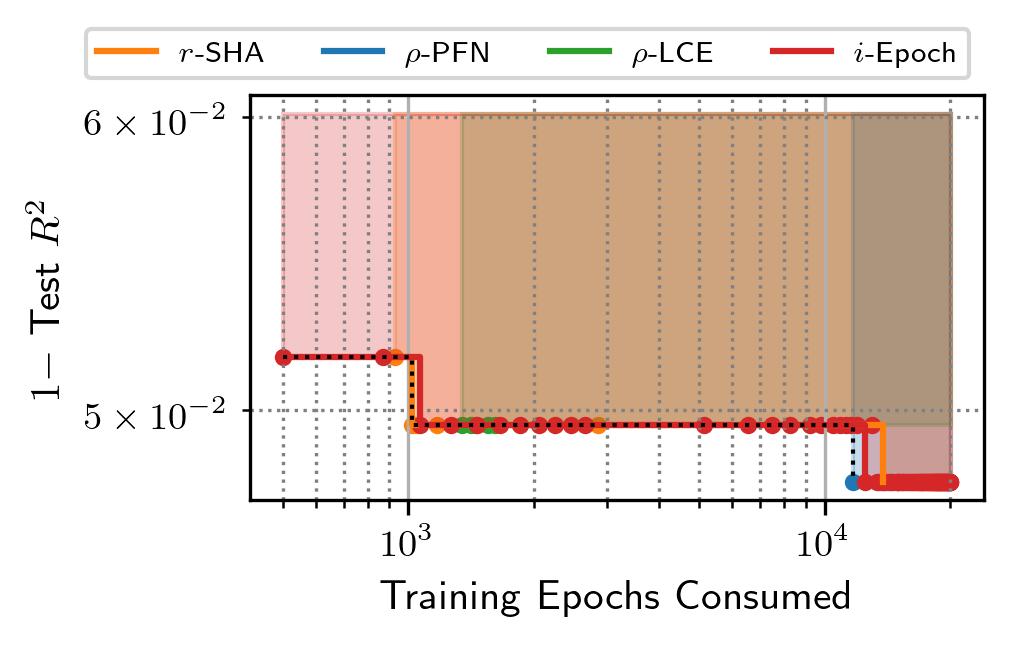}
        \caption{MNIST}
        \label{fig:paretofront-mfhpo-mnist}
    \end{subfigure}
    \begin{subfigure}[b]{0.32\textwidth}
        \centering
        \includegraphics[width=\textwidth]{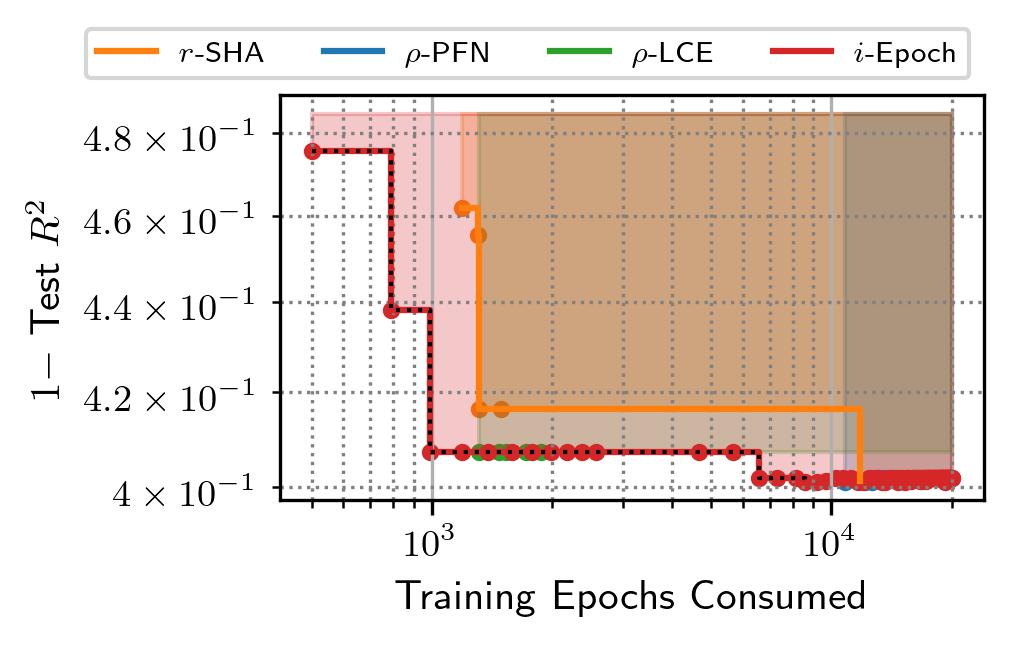}
        \caption{Australian Electricity Market}
        \label{fig:paretofront-mfhpo-electricitymarket}
    \end{subfigure}
    \begin{subfigure}[b]{0.32\textwidth}
        \centering
        \includegraphics[width=\textwidth]{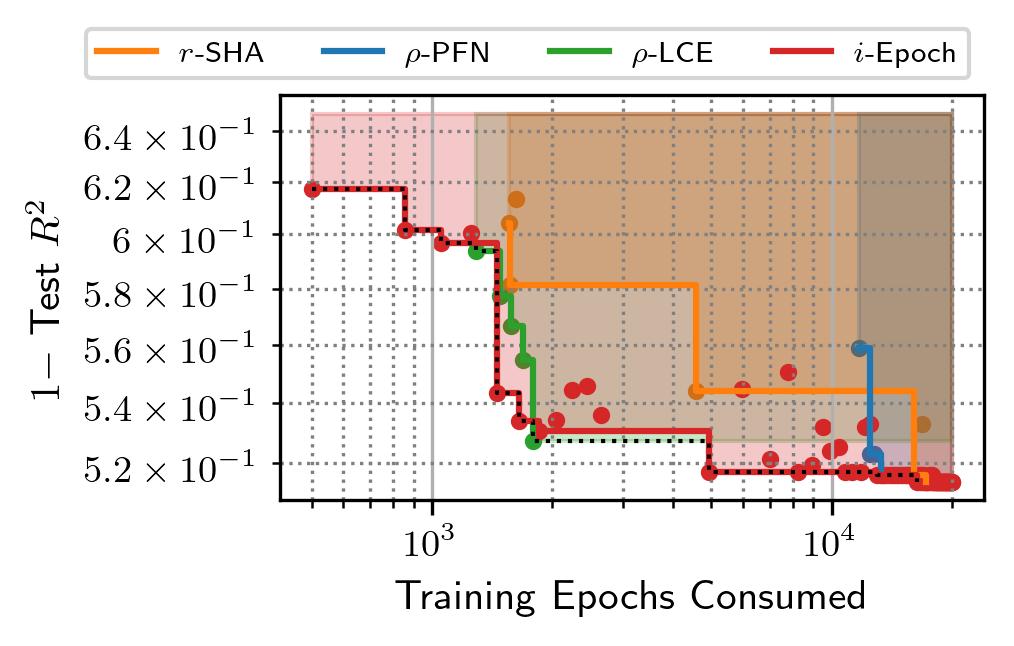}
        \caption{Bank Marketing}
        \label{fig:paretofront-mfhpo-bankmarketing}
    \end{subfigure}
    \begin{subfigure}[b]{0.32\textwidth}
        \centering
        \includegraphics[width=\textwidth]{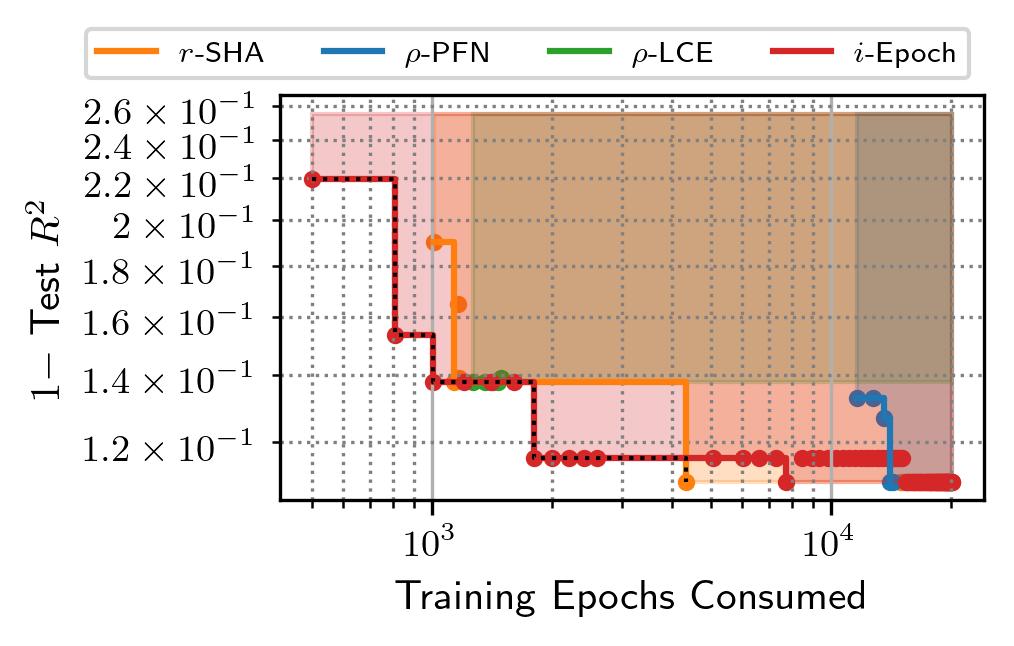}
        \caption{Letter Recognition}
        \label{fig:paretofront-mfhpo-letterrecognition}
    \end{subfigure}
    \begin{subfigure}[b]{0.32\textwidth}
        \centering
        \includegraphics[width=\textwidth]{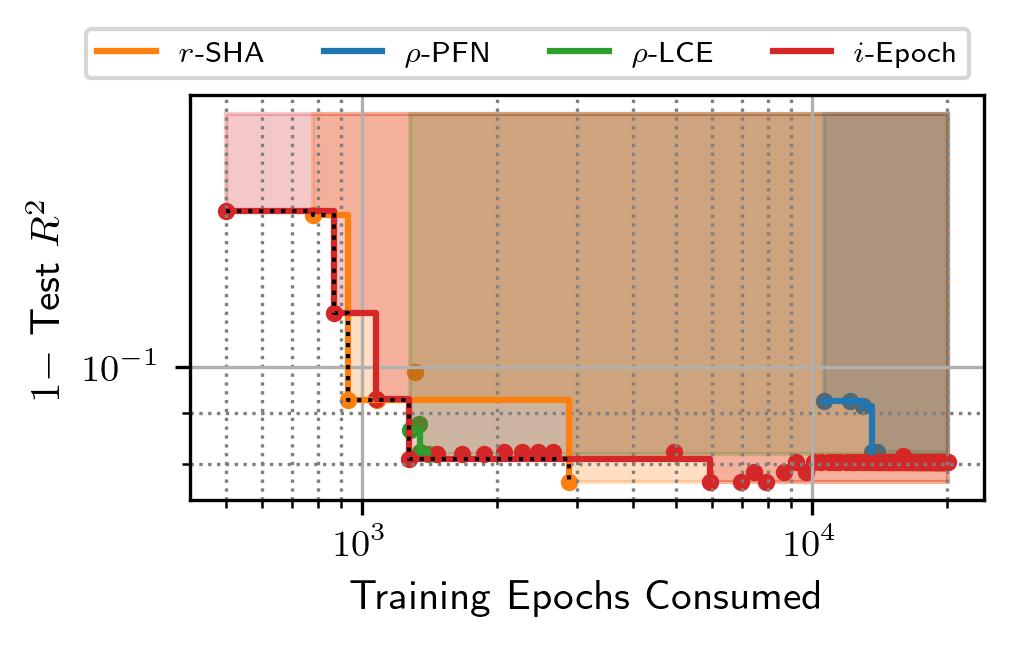}
        \caption{Letter Speech Recognition}
        \label{fig:paretofront-mfhpo-speechrecognition}
    \end{subfigure}
    \begin{subfigure}[b]{0.32\textwidth}
        \centering
        \includegraphics[width=\textwidth]{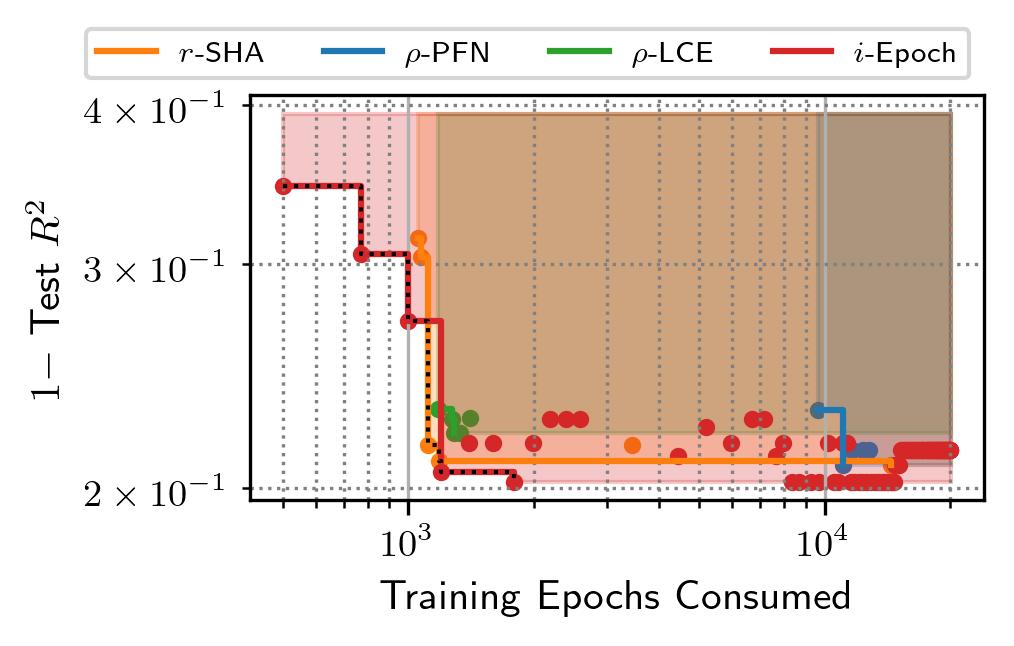}
        \caption{Robot Navigation}
        \label{fig:paretofront-mfhpo-robotnavigation}
    \end{subfigure}
    \begin{subfigure}[b]{0.32\textwidth}
        \centering
        \includegraphics[width=\textwidth]{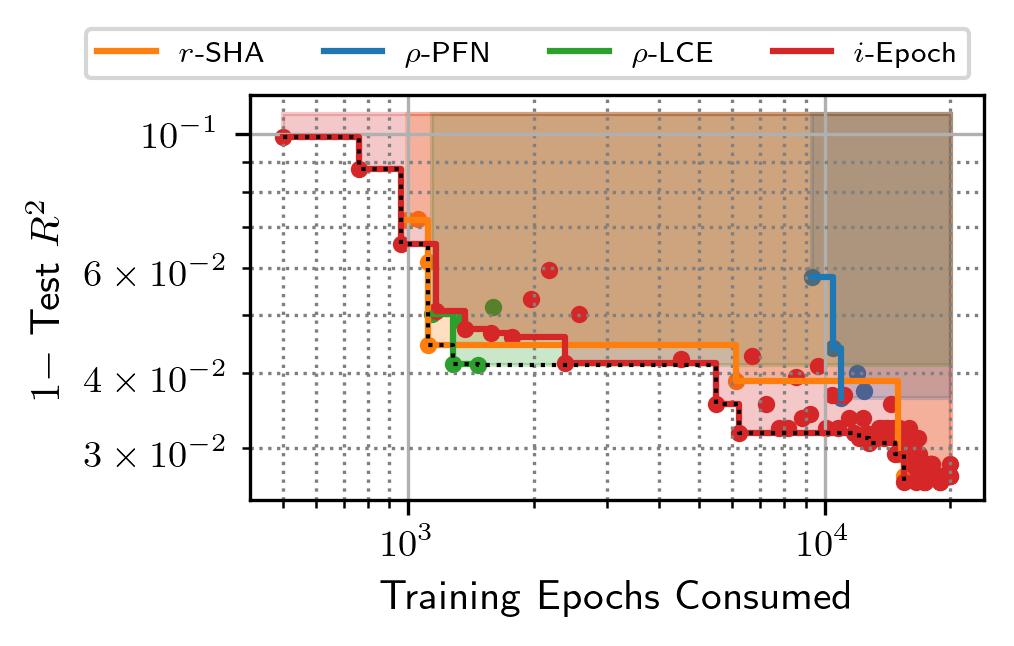}
        \caption{Chess End-Game}
        \label{fig:paretofront-mfhpo-chessendgame}
    \end{subfigure}
    \begin{subfigure}[b]{0.32\textwidth}
        \centering
        \includegraphics[width=\textwidth]{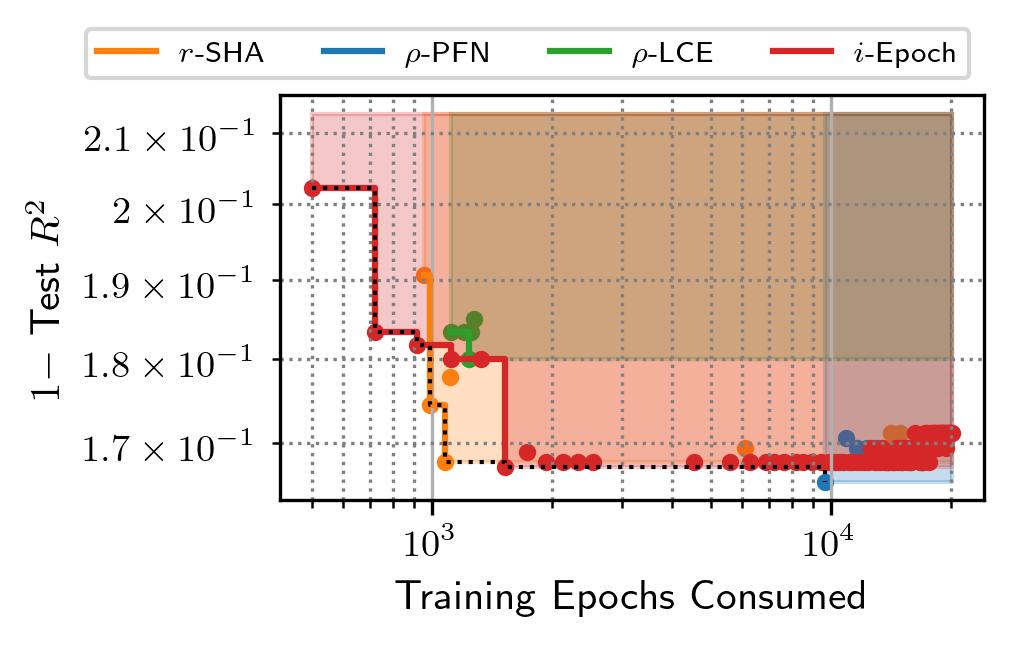}
        \caption{Multiple Features (Karhunen)}
        \label{fig:paretofront-mfhpo-mfkarhunen}
    \end{subfigure}
        \begin{subfigure}[b]{0.32\textwidth}
        \centering
        \includegraphics[width=\textwidth]{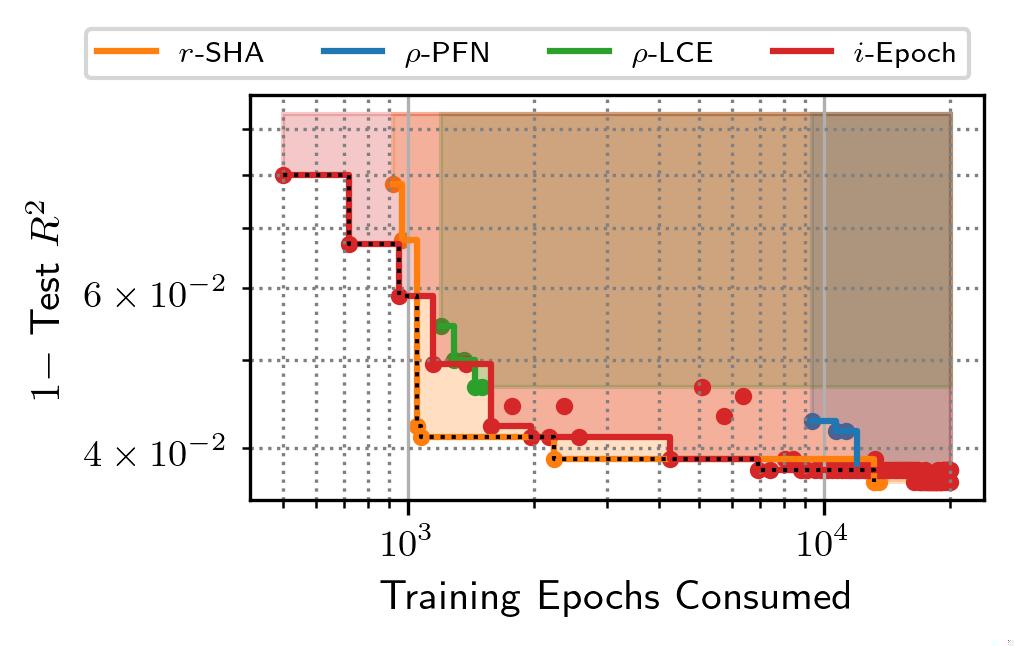}
        \caption{Multiple Features (Fourier)}
        \label{fig:paretofront-mfhpo-mffourier}
    \end{subfigure}
    \begin{subfigure}[b]{0.32\textwidth}
        \centering
        \includegraphics[width=\textwidth]{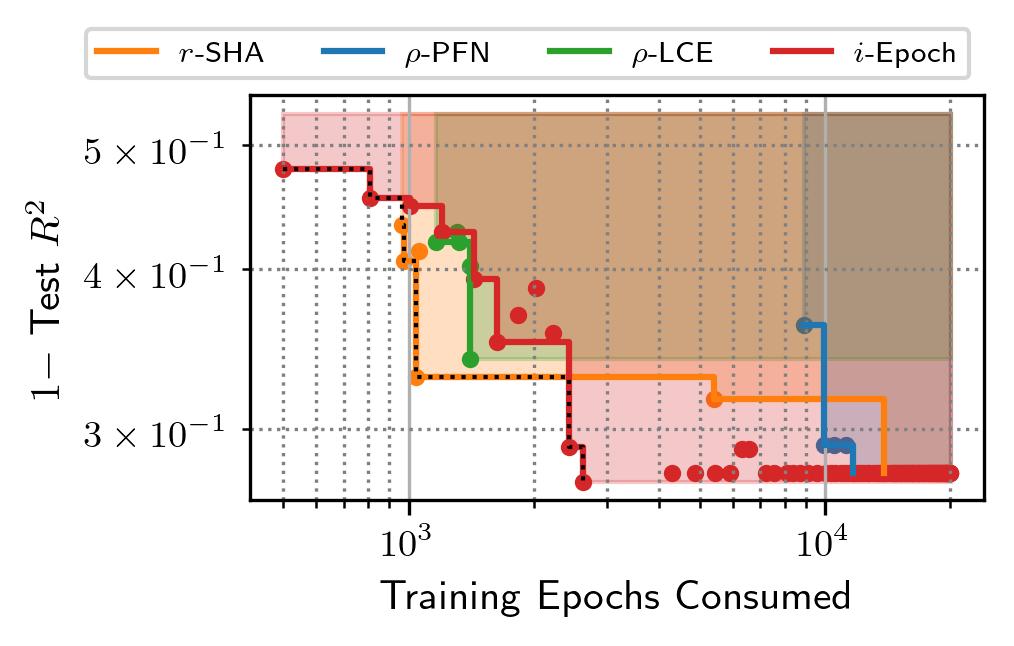}
        \caption{Steel Plates Faults}
        \label{fig:paretofront-mfhpo-steelplatesfaults}
    \end{subfigure}
    \begin{subfigure}[b]{0.32\textwidth}
        \centering
        \includegraphics[width=\textwidth]{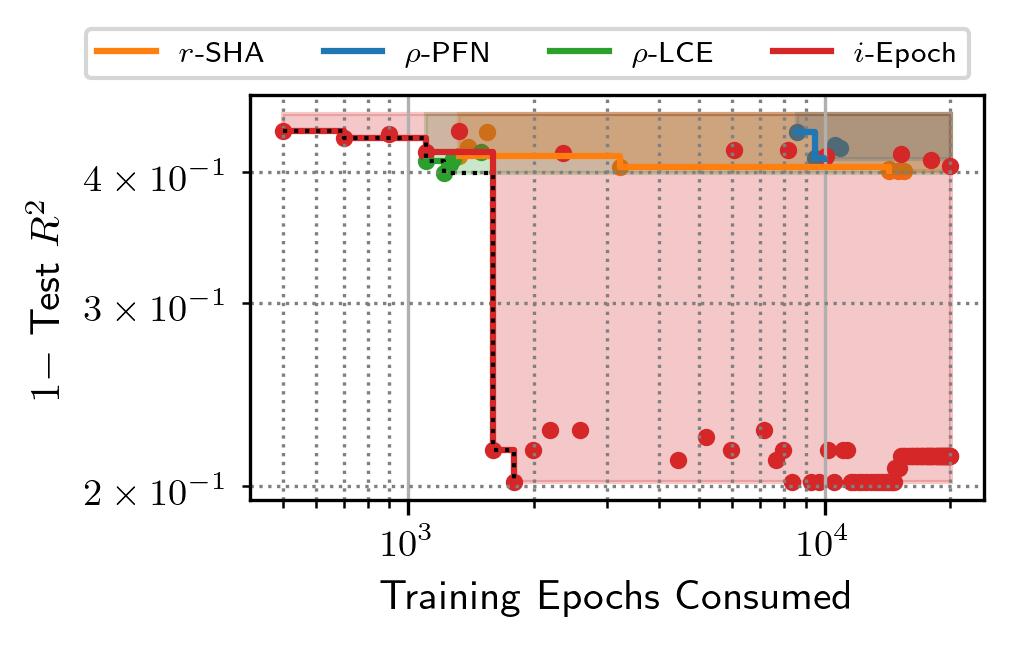}
        \caption{QSAR Biodegradation}
        \label{fig:paretofront-mfhpo-qsarbiodegradation}
    \end{subfigure}
    \begin{subfigure}[b]{0.32\textwidth}
        \centering
        \includegraphics[width=\textwidth]{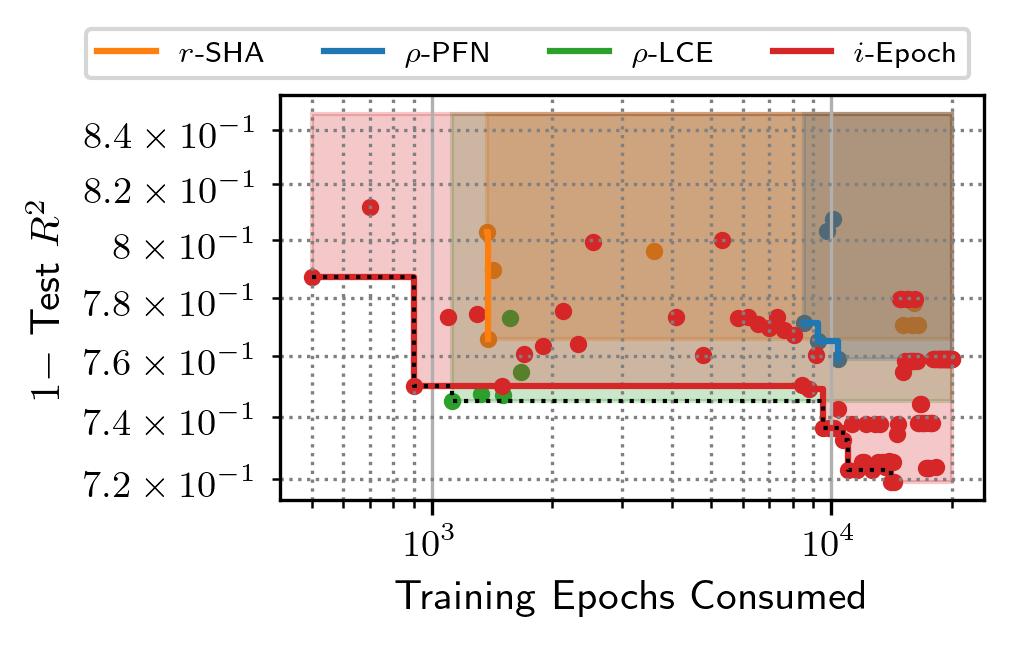}
        \caption{German Credit}
        \label{fig:paretofront-mfhpo-germancredit}
    \end{subfigure}
        \begin{subfigure}[b]{0.32\textwidth}
        \centering
        \includegraphics[width=\textwidth]{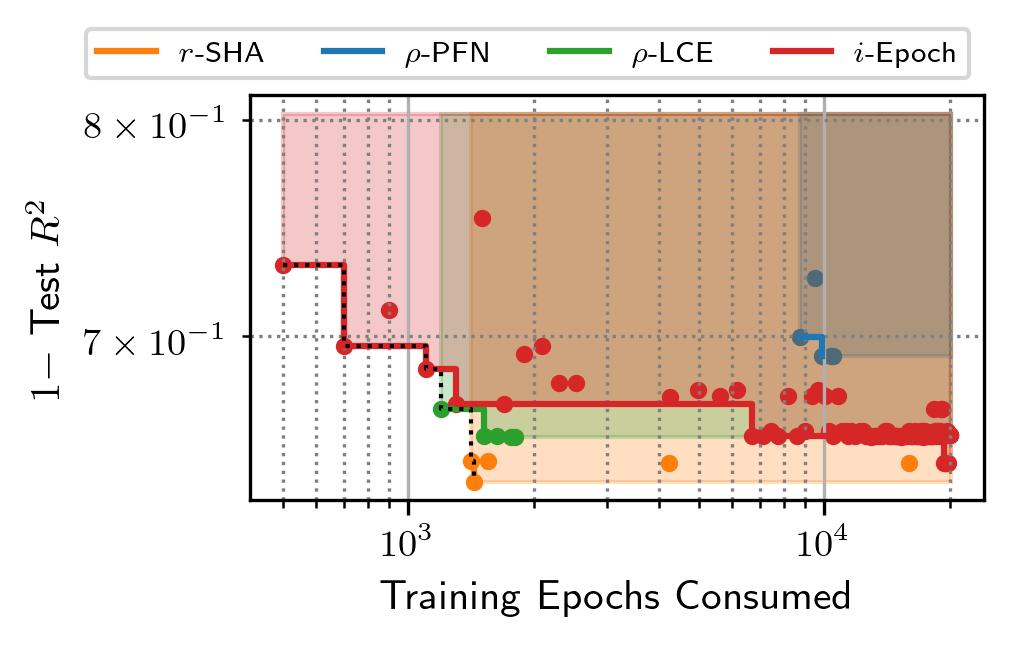}
        \caption{Blood Transfusion}
        \label{fig:paretofront-mfhpo-bloodtransfusion}
    \end{subfigure}
    \caption{Multi-objective profiles built from spanning various levels of aggressiveness of early discarding methods (13 classification tasks). The estimated Pareto-Front of each method is shown in a plane line. The black dotted line corresponds to the estimated Pareto-Front including the methods altogether. {\bf It can be seen that the $i$-Epoch strategy spans more trade-offs (larger area) than other methods while never being significantly dominated}.
    \vspace{-5em}}
    \label{fig:paretofront-mfhpo-classification}
\end{figure}

While the previous question only considers two extreme configurations to understand the sensitivity of the HPO process with respect to the aggressiveness of the early discarding technique, we now want to better understand the actual trade-offs that each method can span.
At this point, we no longer look at any-time performance but instead, we look at the final predictive performance and overall consumed training epochs for one aggressiveness setting.
Once all methods and all aggressiveness levels are collected we compute the Pareto-Front of each early discarding method which does not always contain all evaluated points.

From the results presented in the previous section, we already know that the Pareto-Fronts of $\rho$-PFN will be strictly dominated by other techniques (i.e., the area/hypervolume it defines will be strictly included in the area of other methods).
Since even the difference between minimum ($\rho=0.95$) and maximum aggressiveness ($\rho=0.5$) had only minimal effect, one expects the area covered by $\rho$-PFN in the multi-objective profile to be narrow.

The multi-objective profiles and the corresponding Pareto-Fronts are presented in Figures~\ref{fig:paretofront-mfhpo-regression} and \ref{fig:paretofront-mfhpo-classification}.
For $i$-Epoch, a value was computed for each $1 \leq i \leq 100$.
For $\rho$-LCE and $\rho$-PFN, we used values of $\rho \in \{0.5, 0.7, 0.8, 0.9, 0.95\}$, and for $r$-SHA we used values of $r \in \{\sqrt{\sqrt{2}}=1.19, \sqrt{2}=1.41, 2, 4, 8, 16, 32, 64\}$.
For each approach, the Pareto-optimal points are connected with a step function to indicate the respective Pareto frontier.
The shaded areas show the hypervolume of each approach.

Some plots, like in Figure~\ref{fig:paretofront-mfhpo-navalpropulsion}, suggest a certain inconsistency in the trade-off logic of $i$-Epoch in the sense that many points of a single method do not lie on the same method's Pareto frontier.
However, this can often be attributed to noise on rather small scales.
For example, in the mentioned plot, differences are on a scale below $10^{-3}$, i.e., less than $0.1\%$ difference in performance in terms of the constant predictor baseline.
For the other methods, this effect is less pronounced or does not occur because much fewer points are generated and the change in aggressiveness is more significant ($10\%$-steps in the case of $\rho$ compared to single epochs in the case of $i$-Epoch).

The first observation confirms our expectation that learning curve extrapolation-based techniques offer little diversity of trade-offs.
$\rho$-LCE, no matter how aggressiveness is configured, tends to use about 10x less training epoch than 100-Epoch while sometimes slightly under-performing in attained predictive performance.
And, again, PFN on most datasets offers almost no reductions regardless of the configuration of $\rho$.

\FloatBarrier

\subsection{RQ3 -- Which methods offer diverse trade-offs?}

To quantity the observation that $i$-Epoch offers a more diverse set of trade-offs we compute the relative hypervolume spanned by each method in Figures~\ref{fig:paretofront-mfhpo-regression} and ~\ref{fig:paretofront-mfhpo-classification}.
To evaluate the hypervolume we set as reference point $y_\text{ref} := (\max \mu_L + \sigma^\text{err}_L , \max \mu_B + \sigma^\text{err}_B)$ (i.e., element-wise upper-bound of observations) for all methods. 
Then we apply a $\log_{10}(.)$ transformations on both $y_L$ and $y_I$ values (including the reference point). 
This transformation serves to spread the volume contributed by small and large values equally.
Otherwise, differences in hypervolume would become unnoticeable as soon as improvements in $y_L$ or $y_I$ become orders of magnitude smaller than the largest reference point values.
Finally, we compute the hypervolume of all methods which we divide the hypervolume of the Pareto-Front considering all observations (in dotted black line).
This relative hypervolume then quantifies how much each method contributes to the available set of trade-offs that we observed.
The closer is the value to 1 the more complete the method.
The resulting scores are presented in Table~\ref{tab:hvi-scores}. 

As it can be observed $i$-Epoch achieves the highest scores on all but one task giving it an average rank of 1.125. The second best-ranked method is $\rho$-SHA followed by $\rho$-LCE. The $\rho$-PFN method consistently finishes last ranked on all tested tasks. Lastly, we also notice that {\bf relative hypervolume scores of $i$-Epoch are often close to 1 which confirms that this method spans most of the observed trade-offs and it is never significantly outperformed in either objective}.

\begin{table}[!h]
\begin{tabular}{ccccc}
\hline
\textbf{Dataset}              & \multicolumn{1}{l}{\textbf{$r$-SHA}}                 & \multicolumn{1}{l}{\textbf{$\rho$-PFN}}              & \multicolumn{1}{l}{\textbf{$\rho$-LCE}}              & \multicolumn{1}{l}{\textbf{$i$-Epoch}}                        \\ \hline
Slice Localization            & {\color[HTML]{E7B416} 0.930}                         & {\color[HTML]{CC3232} 0.401}                         & {\color[HTML]{DB7B2B} 0.823}                         & {\color[HTML]{2DC937} \textbf{0.932}}                         \\
Protein Structure             & {\color[HTML]{E7B416} 0.916}                         & {\color[HTML]{CC3232} 0.241}                         & {\color[HTML]{DB7B2B} 0.652}                         & {\color[HTML]{2DC937} \textbf{0.989}}                         \\
Naval Propulsion              & {\color[HTML]{E7B416} 0.881}                         & {\color[HTML]{CC3232} 0.280}                         & {\color[HTML]{DB7B2B} 0.742}                         & {\color[HTML]{2DC937} \textbf{0.951}}                         \\
Parkinson's Telemonitoring    & {\color[HTML]{2DC937} \textbf{0.930}}                & {\color[HTML]{CC3232} 0.201}                         & {\color[HTML]{DB7B2B} 0.667}                         & {\color[HTML]{E7B416} 0.880}                                  \\ \hline
MNIST                         & {\color[HTML]{E7B416} 0.858}                         & {\color[HTML]{CC3232} 0.176}                         & {\color[HTML]{DB7B2B} 0.743}                         & {\color[HTML]{2DC937} \textbf{0.994}}                         \\
Australian Electricity Market & {\color[HTML]{DB7B2B} 0.768}                         & {\color[HTML]{CC3232} 0.205}                         & {\color[HTML]{E7B416} 0.829}                         & {\color[HTML]{2DC937} \textbf{1.000}}                         \\
Bank Marketing                & {\color[HTML]{DB7B2B} 0.609}                         & {\color[HTML]{CC3232} 0.184}                         & {\color[HTML]{E7B416} 0.847}                         & {\color[HTML]{2DC937} \textbf{0.989}}                         \\
Letter Recognition            & {\color[HTML]{E7B416} 0.851}                         & {\color[HTML]{CC3232} 0.169}                         & {\color[HTML]{F8A102} 0.672}                         & {\color[HTML]{2DC937} \textbf{0.988}}                         \\
Letter Speech Recognition     & {\color[HTML]{E7B416} 0.915}                         & {\color[HTML]{CC3232} 0.175}                         & {\color[HTML]{F8A102} 0.810}                         & {\color[HTML]{2DC937} \textbf{0.974}}                         \\
Robot Navigation              & {\color[HTML]{E7B416} 0.882}                         & {\color[HTML]{CC3232} 0.218}                         & {\color[HTML]{F8A102} 0.789}                         & {\color[HTML]{2DC937} \textbf{0.992}}                         \\
Chess End-Game                & {\color[HTML]{E7B416} 0.866}                         & {\color[HTML]{CC3232} 0.233}                         & {\color[HTML]{F8A102} 0.827}                         & {\color[HTML]{2DC937} \textbf{0.965}}                         \\
Multiple Features (Karhunen)  & {\color[HTML]{E7B416} 0.901}                         & {\color[HTML]{CC3232} 0.231}                         & {\color[HTML]{F8A102} 0.606}                         & {\color[HTML]{2DC937} \textbf{0.955}}                         \\
Multiple Features (Fourier)   & {\color[HTML]{E7B416} 0.936}                         & {\color[HTML]{CC3232} 0.239}                         & {\color[HTML]{F8A102} 0.697}                         & {\color[HTML]{2DC937} \textbf{0.951}}                         \\
Steel Plates Faults           & {\color[HTML]{E7B416} 0.806}                         & {\color[HTML]{CC3232} 0.257}                         & {\color[HTML]{F8A102} 0.644}                         & {\color[HTML]{2DC937} \textbf{0.923}}                         \\
QSAR Biodegradation           & {\color[HTML]{DB7B2B} 0.141}                         & {\color[HTML]{CC3232} 0.037}                         & {\color[HTML]{E7B416} 0.176}                         & {\color[HTML]{2DC937} \textbf{0.993}}                         \\
German Credit                 & {\color[HTML]{DB7B2B} 0.585}                         & {\color[HTML]{CC3232} 0.198}                         & {\color[HTML]{E7B416} 0.800}                         & {\color[HTML]{2DC937} \textbf{0.970}}                         \\
Blood Transfusion             & {\color[HTML]{E7B416} 0.811}                         & {\color[HTML]{CC3232} 0.167}                         & {\color[HTML]{DB7B2B} 0.753}                         & {\color[HTML]{2DC937} \textbf{0.856}}                         \\ \hline
\textbf{Average Rank}         & \cellcolor[HTML]{F2F2F2}{\color[HTML]{E7B416} 2.029} & \cellcolor[HTML]{F2F2F2}{\color[HTML]{CC3232} 4.000} & \cellcolor[HTML]{F2F2F2}{\color[HTML]{DB7B2B} 2.846} & \cellcolor[HTML]{F2F2F2}{\color[HTML]{2DC937} \textbf{1.125}} \\ \hline
\end{tabular}
\caption{Relative hypervolumes of each early discarding technique with respect to the hypervolume including all the techniques. {\color[HTML]{2DC937}Bold and green} is best, followed by {\color[HTML]{E7B416}yellow}, {\color[HTML]{DB7B2B}orange} and {\color[HTML]{CC3232}red}. These scores assess the diversity of trade-offs, in consumed training epochs and predictive performance, offered by each technique among all observed outcomes. The higher the score the more complete (in terms of possible trade-offs) is the early discarding technique. In our experiments, {\bf the $i$-Epoch technique offers the best set of trade-offs} and achieves a trade-off close to 1 indicating optimality amongst all methods.}
\label{tab:hvi-scores}
\end{table}

\subsection{RQ4 -- Is 1-Epoch so good and if so, how?}
\label{sec:aggressive}

Last but not least, throughout our presented results we can notice the unreasonable effectiveness of 1-Epoch. Despite sometimes being noisier in its performance profiles such as in Figures~\ref{fig:curves-mfhpo-navalpropulsion}, \ref{fig:paretofront-mfhpo-robotnavigation} and,\ref{fig:paretofront-mfhpo-germancredit}, it always achieved better any-time performance than other early discarding methods.
This is demonstrated by the fact that its performance curve does not cross with the performance curves of other methods.
However, the difference in final predictive performance $y_L$ can sometimes be statistically significant such as in Figures~\ref{fig:curves-mfhpo-bankmarketing}, \ref{fig:curves-mfhpo-chessendgame}, \ref{fig:curves-mfhpo-mffourier} and, \ref{fig:curves-mfhpo-steelplatesfaults} which confirms the trade-off between the two objectives.

\begin{figure}[t]
    \centering
    \begin{subfigure}[b]{0.49\textwidth}
        \centering
        \includegraphics[width=\textwidth]{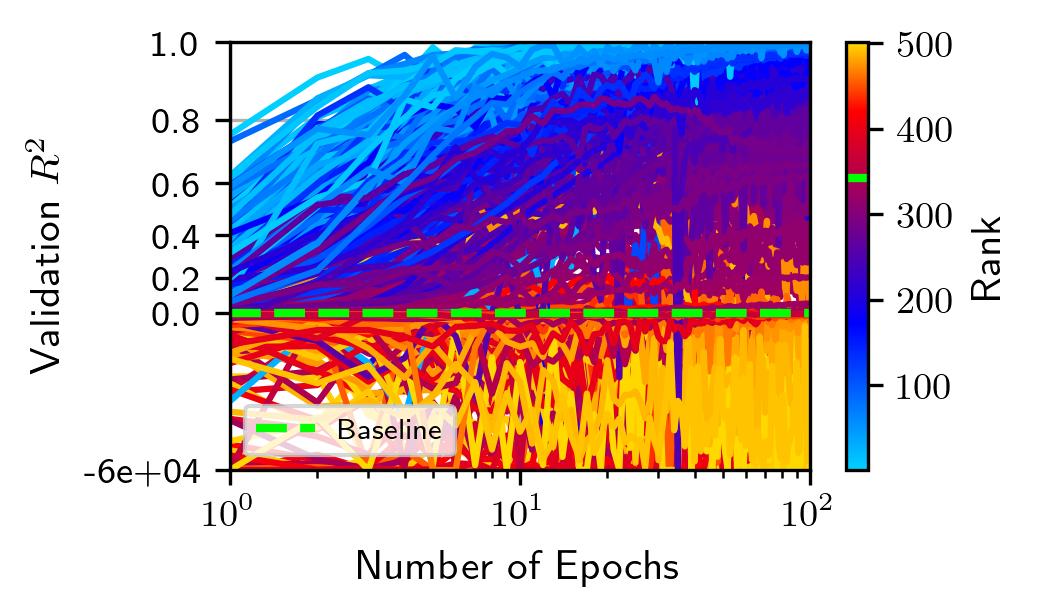}
        \caption{Naval Propulsion}
        \label{fig:viz-learning-curves-navalpropulsion}
    \end{subfigure}
    \begin{subfigure}[b]{0.49\textwidth}
        \centering
        \includegraphics[width=\textwidth]{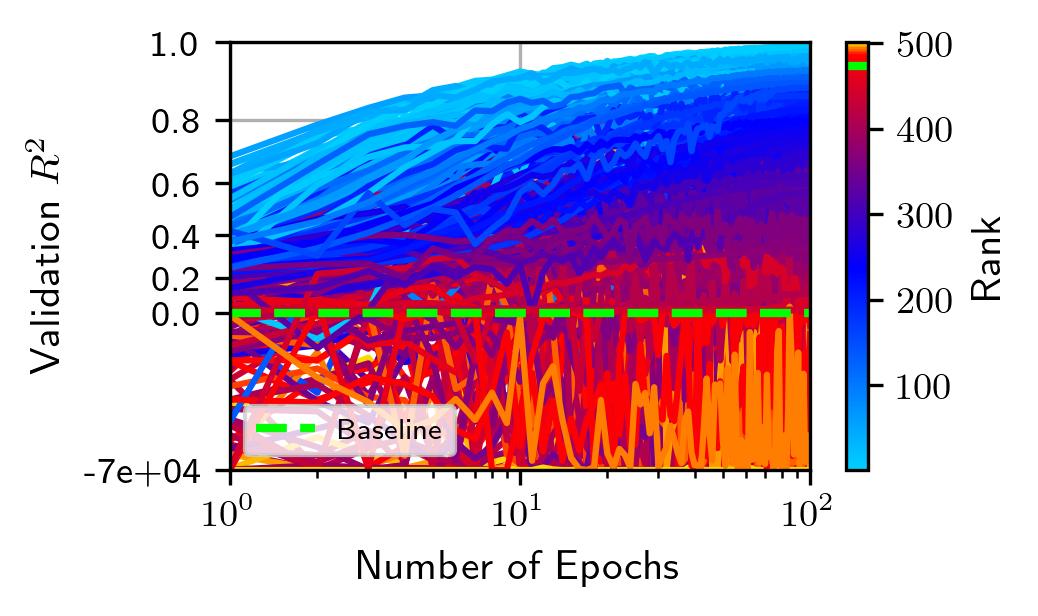}
        \caption{Parkinson's Telemonitoring}
        \label{fig:viz-learning-curves-parkinsons}
    \end{subfigure}\\
    \begin{subfigure}[b]{0.49\textwidth}
        \centering
        \includegraphics[width=\textwidth]{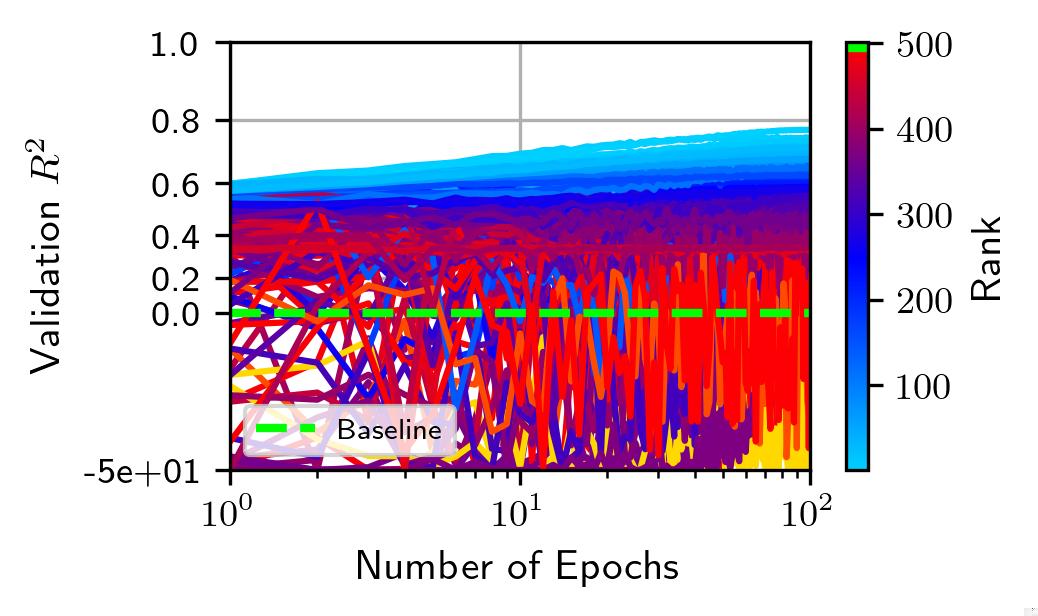}
        \caption{Protein Structure}
        \label{fig:viz-learning-curves-protein}
    \end{subfigure}
    \begin{subfigure}[b]{0.49\textwidth}
        \centering
        \includegraphics[width=\textwidth]{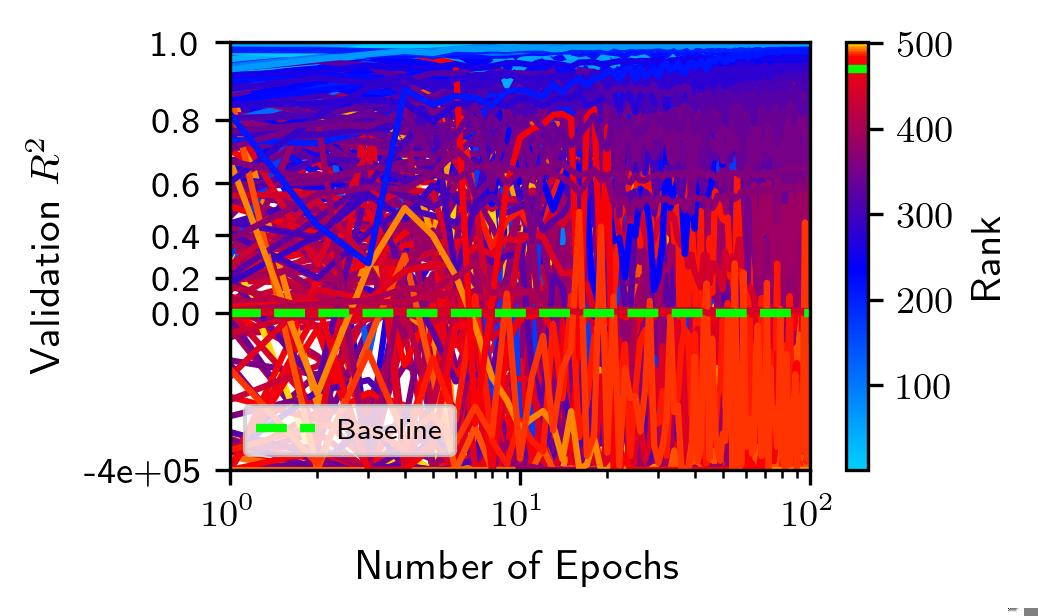}
        \caption{Slice Localization}
        \label{fig:viz-learning-curves-slicelocalization}
    \end{subfigure}
    \caption{Visualizing the final ranking for {\color{blueplot}good (light blue)} and {\color{yellowplot}bad (yellow)} models for 500 randomly sampled learning curves (on 4 regression tasks). The constant predictor performance (at 0) is shown as a green dashed line. {\bf Models can be selected from the first epoch as there appear to be dominant models early on in the training epochs}. }
    \label{fig:viz-learning-curves-regression}
\end{figure}

\begin{figure}[t]
    \centering
    \begin{subfigure}[b]{0.32\textwidth}
        \centering
        \includegraphics[width=\textwidth]{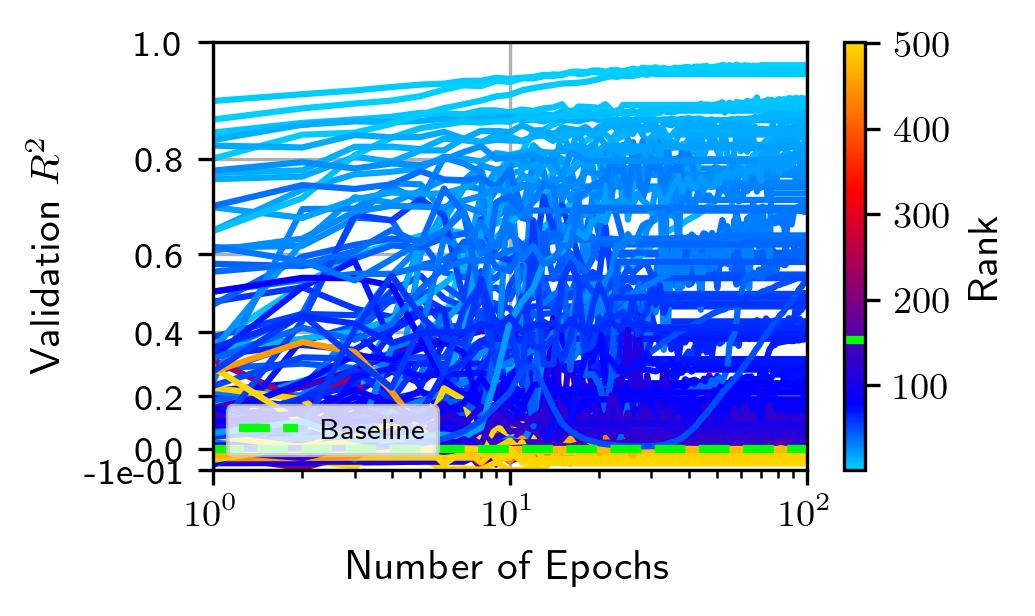}
        \caption{MNIST}
        \label{fig:viz-learning-curves-mnist}
    \end{subfigure}
    \begin{subfigure}[b]{0.32\textwidth}
        \centering
        \includegraphics[width=\textwidth]{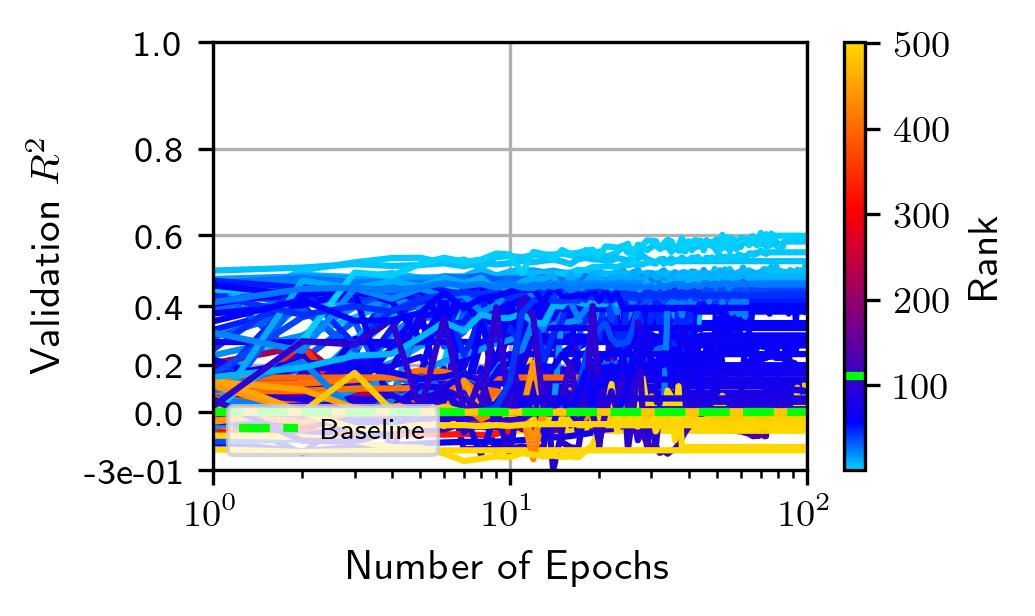}
        \caption{Australian Electricity Market}
        \label{fig:viz-learning-curves-electricitymarket}
    \end{subfigure}
    \begin{subfigure}[b]{0.32\textwidth}
        \centering
        \includegraphics[width=\textwidth]{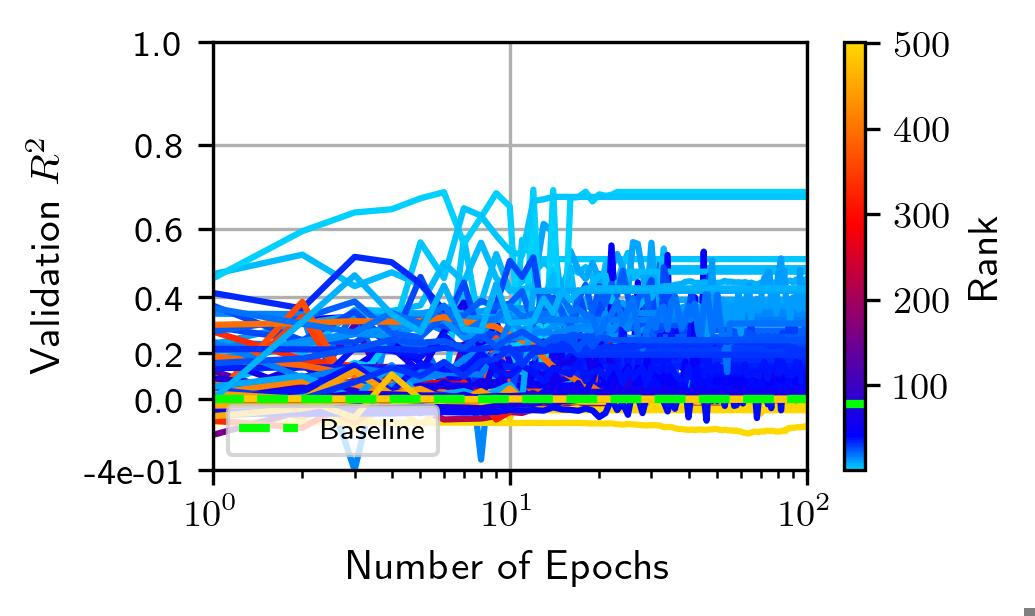}
        \caption{Bank Marketing}
        \label{fig:viz-learning-curves-bankmarketing}
    \end{subfigure}
    \begin{subfigure}[b]{0.32\textwidth}
        \centering
        \includegraphics[width=\textwidth]{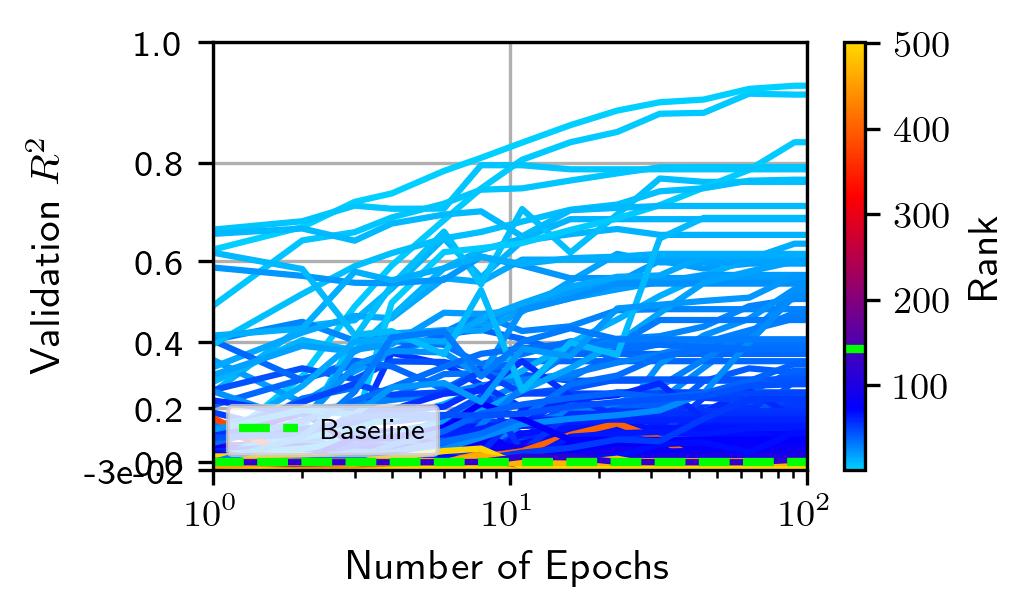}
        \caption{Letter Recognition}
        \label{fig:viz-learning-curves-letterrecognition}
    \end{subfigure}
    \begin{subfigure}[b]{0.32\textwidth}
        \centering
        \includegraphics[width=\textwidth]{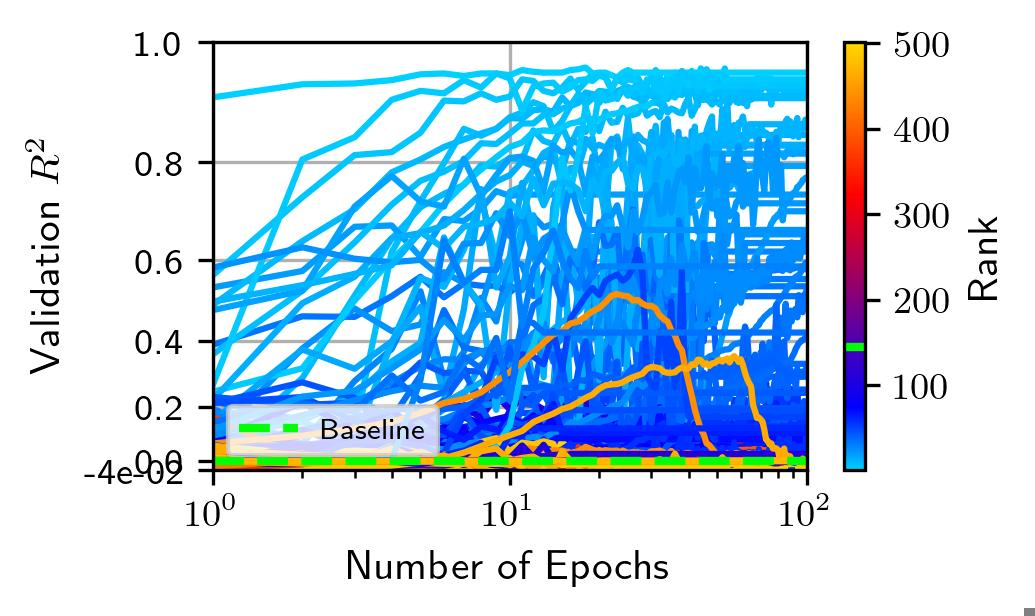}
        \caption{Letter Speech Recognition}
        \label{fig:viz-learning-curves-speechrecognition}
    \end{subfigure}
    \begin{subfigure}[b]{0.32\textwidth}
        \centering
        \includegraphics[width=\textwidth]{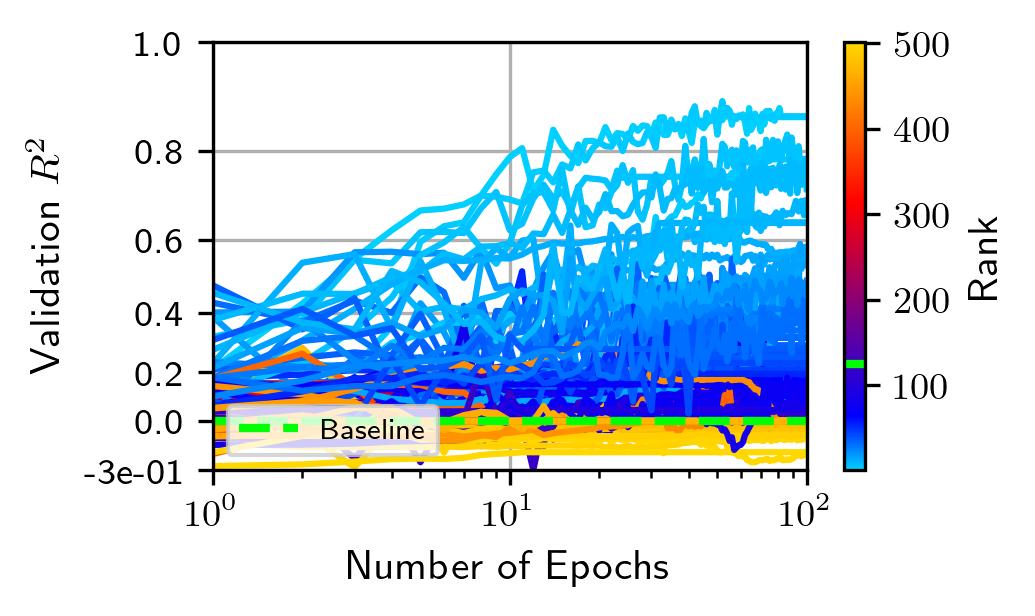}
        \caption{Robot Navigation}
        \label{fig:viz-learning-curves-robotnavigation}
    \end{subfigure}
    \begin{subfigure}[b]{0.32\textwidth}
        \centering
        \includegraphics[width=\textwidth]{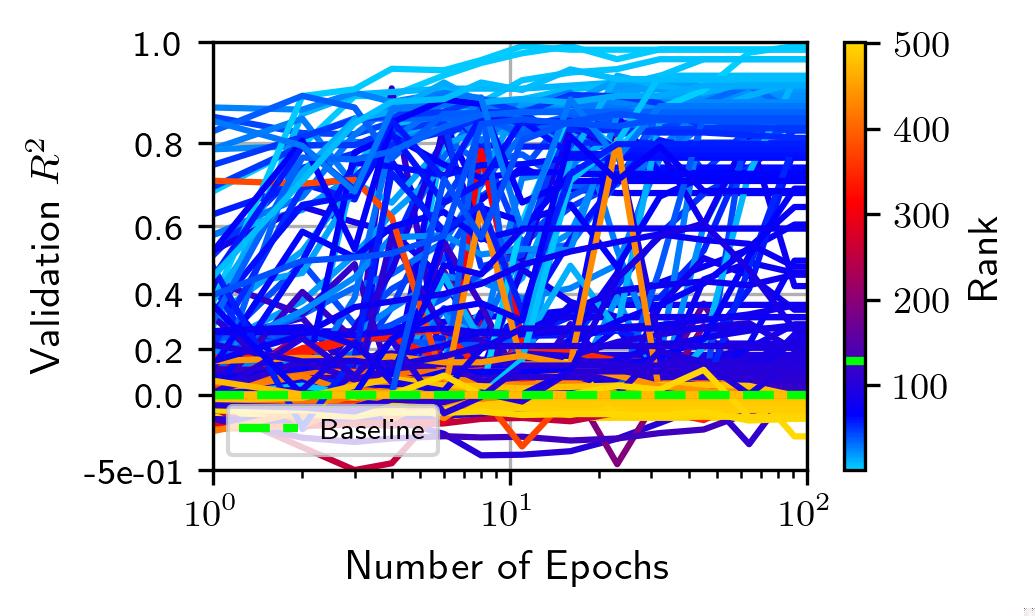}
        \caption{Chess End-Game}
        \label{fig:viz-learning-curves-chessendgame}
    \end{subfigure}
    \begin{subfigure}[b]{0.32\textwidth}
        \centering
        \includegraphics[width=\textwidth]{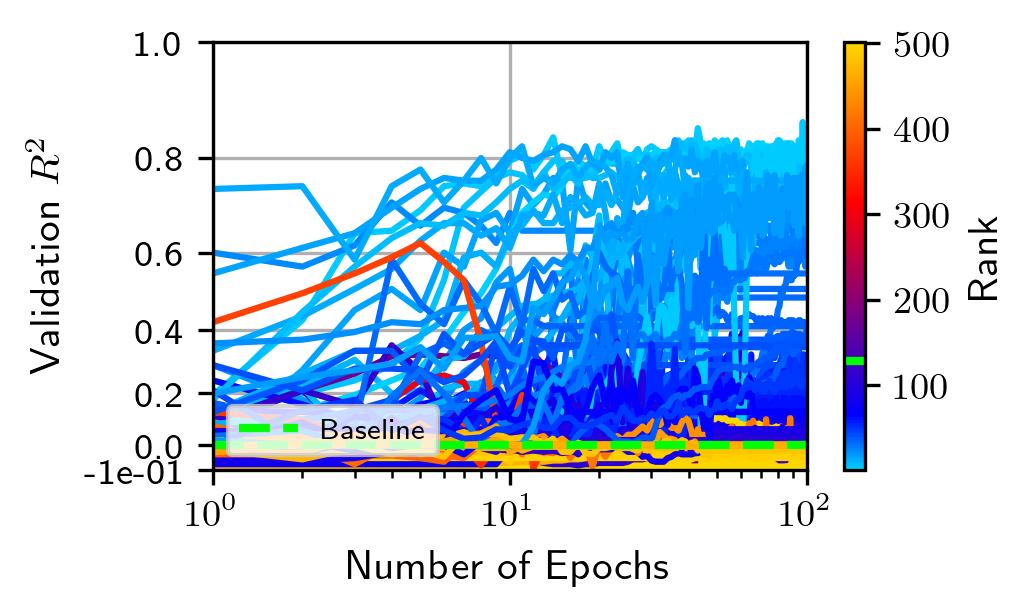}
        \caption{Multiple Features (Karhunen)}
        \label{fig:viz-learning-curves-mfkarhunen}
    \end{subfigure}
    \begin{subfigure}[b]{0.32\textwidth}
        \centering
        \includegraphics[width=\textwidth]{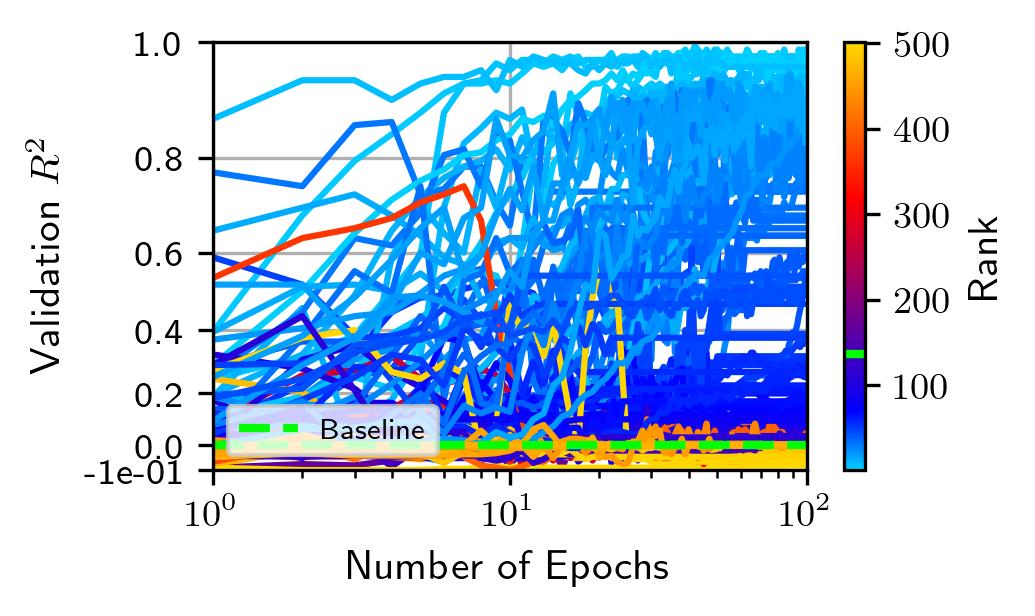}
        \caption{Multiple Features (Fourier)}
        \label{fig:viz-learning-curves-mffourier}
    \end{subfigure}
    \begin{subfigure}[b]{0.32\textwidth}
        \centering
        \includegraphics[width=\textwidth]{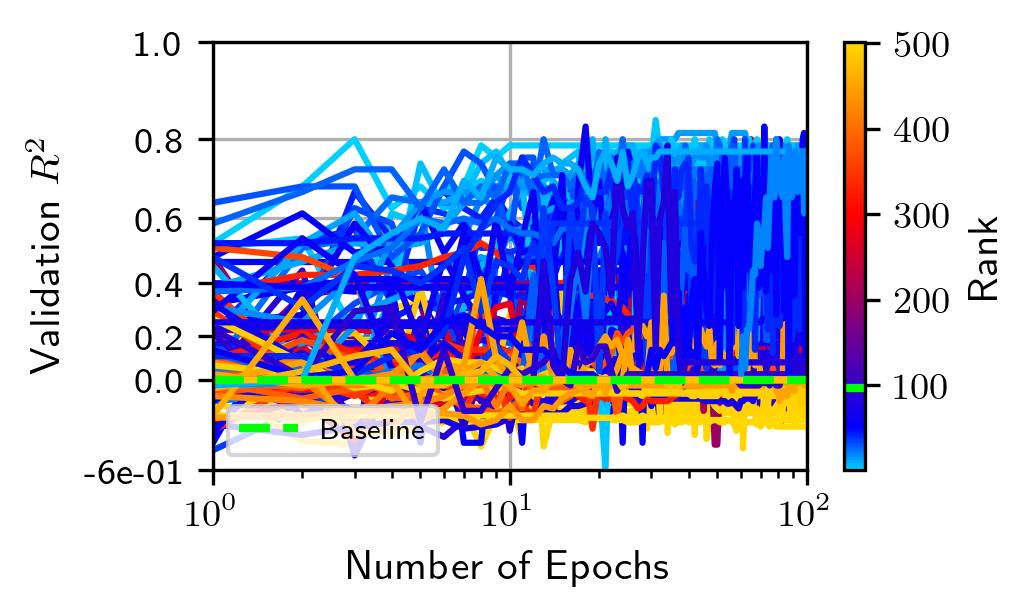}
        \caption{QSAR Biodegradation}
        \label{fig:viz-learning-curves-qsarbiodegradation}
    \end{subfigure}
    \begin{subfigure}[b]{0.32\textwidth}
        \centering
        \includegraphics[width=\textwidth]{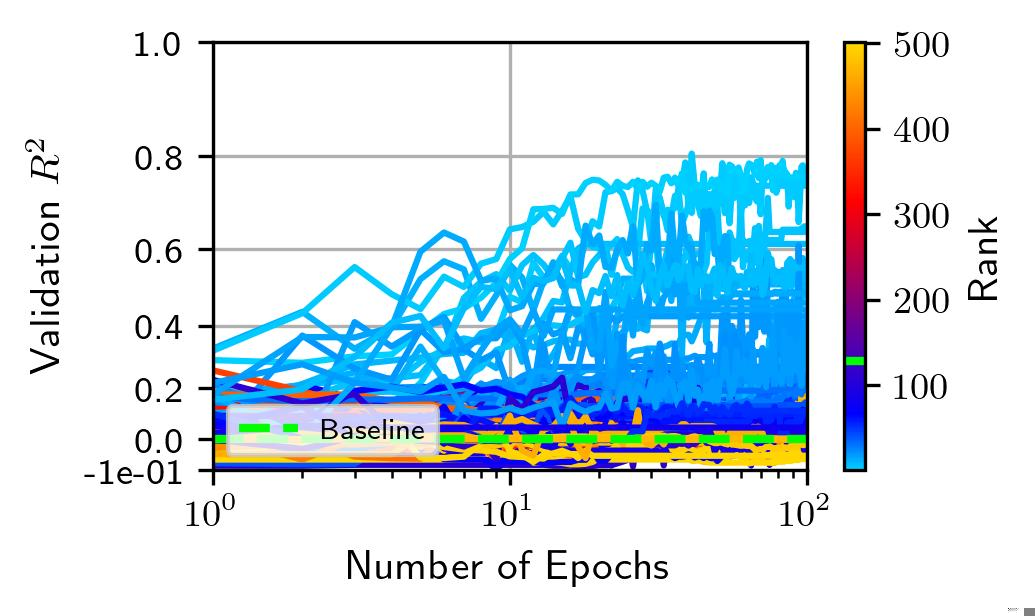}
        \caption{Steel Plates Faults}
        \label{fig:viz-learning-curves-steelplatesfaults}
    \end{subfigure}
    \begin{subfigure}[b]{0.32\textwidth}
        \centering
        \includegraphics[width=\textwidth]{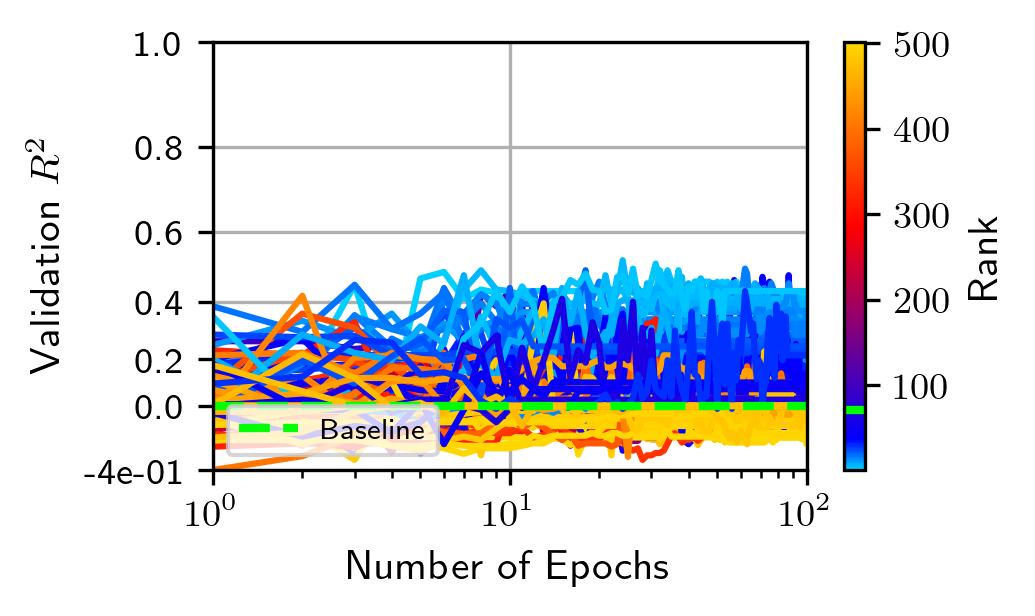}
        \caption{German Credit}
        \label{fig:viz-learning-curves-germancredit}
    \end{subfigure}
    \begin{subfigure}[b]{0.32\textwidth}
        \centering
        \includegraphics[width=\textwidth]{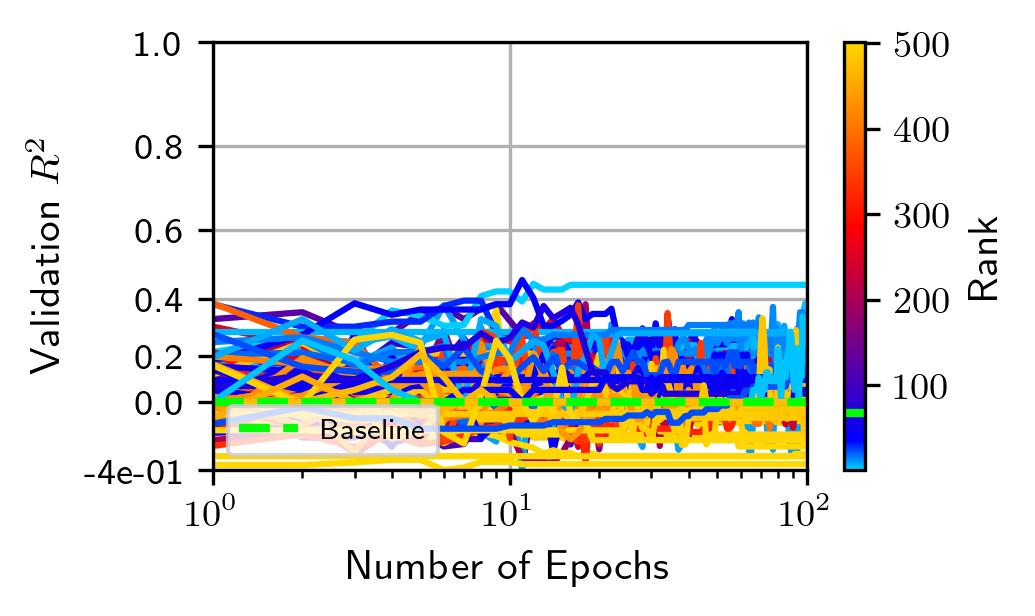}
        \caption{Blood Transfusion}
        \label{fig:viz-learning-curves-bloodtransfusion}
    \end{subfigure}
    \caption{Visualizing the final ranking for {\color{blueplot}good (light blue)} and {\color{yellowplot}bad (yellow)} models for 500 randomly sampled learning curves (on 13 classification tasks). The constant predictor performance (at 0) is shown as a green dashed line. {\bf Models can be selected from the first epoch as there appear to be dominant models early on in the training epochs}.}
    \label{fig:viz-learning-curves-classification}
\end{figure}

How is it possible this approach can perform so well? To better understand this, we analyze the learning curves of our experiments. 
In Figures~\ref{fig:viz-learning-curves-regression} and \ref{fig:viz-learning-curves-classification} we display 500 randomly sampled learning curves from our pre-computed sets, we then color the curves by their ranking at 100 epochs (the maximum number of training epochs). 
Low ranks, colored in light blue, correspond to the best models, while high ranks, colored in red and then yellow, correspond to the worst models. 
We plot the performance of the constant predictor as a dashed lime line and also plot its rank.

In these plots, it can be observed that {\bf for all benchmarks there exists among the best models some that are also the best early in the training process}. This observation explains the performance of $1$-Epoch.
Then, in a few cases, we can observe {\bf a significant proportion of models perform worse than the constant predictor}. It is about 33\% of models in Figure~\ref{fig:viz-learning-curves-navalpropulsion} and about 80\% of models in \ref{fig:viz-learning-curves-chessendgame} to \ref{fig:viz-learning-curves-mffourier}. Finally, it seems that {\bf learning curve oscillations are correlated with the final predictive performance}.
The best models present much less oscillations than the worst models, which justifies high aggressiveness in the early discarding method.

\FloatBarrier

\section{Conclusion}
\label{sec:conclusion}

In this paper, we conducted a comprehensive analysis of early discarding techniques for hyperparameter optimization of fully connected deep neural networks.
Our study rigorously compared an array of advanced techniques and unveiled intriguing findings: (1) {\bf the unreasonable effectiveness of the 1-Epoch strategy}, a straightforward yet previously overlooked baseline method, and
(2) {\bf the Pareto-dominance of the $i$-Epoch strategy despite its simplicity}. 

We attribute the success of this strategy to effectively differentiating between high and low-potential models in the early stages of training. Notably, models with promising prospects exhibit minimal performance oscillations, a pattern consistently observed in widely used benchmarks. These insights not only underscore the importance of incorporating the $i$-Epoch strategy in future benchmark analyses but also highlight the potential necessity of considering the multi-objective problem hidden behind early discarding strategies.
{\bf An early discarding method would bring significant value only if it complements or dominates the $i$-Epoch Pareto-Front}.
Current early discarding approaches only add minimal or no utility in this sense.

Besides its good performance, we believe that 1-Epoch's simplicity is valuable in itself. Besides being easy to implement, before execution, it is easy to predict the number of training epochs consumed by $i$-Epoch for any $i$ when it is not possible for either $\rho$-LCE or $r$-SHA. This makes $i$-Epoch practically attractive.

To be noted, our work is limited to using ``epoch" as iteration units for early discarding. While this is convenient and appealing to conduct studies independent of hardware implementation considerations, practical application settings may require considering wall time or other options as units for early discarding. In particular, since different configurations may have different batch sizes, some configurations could be much faster to train than others. However, comparing wall-clock time is extremely hardware and software implementation dependent. Maybe considering the size of the deep neural network as a third objective of Equation~\ref{eq:moo-optimization-problem} could be an improvement.

We have tried a limited range for the aggressiveness parameters of $\rho$-LCE and $r$-SHA. Their Pareto-Front could be larger and more dominant for a wider range of parameters considered. 
However, values of $\rho < 0.5$ seem relatively strange for $\rho$-LCE because in that case it will be very pessimistic about extrapolated performance and discard models as soon as there is a small probability of under-performing.
$r$-SHA could be more aggressive but it should be noted that our largest reduction factor of 64 corresponds to continuing training only if the model is in the current Top-1.5\% meaning comparing to the single best model after 100 Hyperparameter suggestions and Top-3 of 200. Also, this value is significantly larger than the suggested default value of 4 in the original paper\cite{li_system_2020}.

Also, we studied the early discarding methods in combination with a random search. In other words, HPO is often combined with techniques that suggest candidates through more sophisticated methods, such as Bayesian optimization or portfolio \cite{jamieson2016non,li_hyperband_2018,falkner_bohb_2018,awad_dehb_2021}. However, for such approaches we cannot quantify the computational cost as easily through the number of epochs, as the Bayesian optimization may not be a neural net. Besides that, the comparison becomes more complicated, because the different components (configuration proposer, early discarding technique, etc.) may interact in unexpected ways. 
Therefore, such a comparison is out of the scope.

To come back to the question of the earlier work: is one epoch all you need? We think the answer remains to be seen, in particular, we think that the 1-epoch approach can be even pushed further. During the first epoch, more information is available for making decisions. For example, the loss per batch could be collected. This again forms a curve of performances versus the number of batches processed, which seems conceptually similar to a learning curve. This curve could be extrapolated as well. This will allow us to make potentially better decisions after 1 epoch or even training could be stopped before finishing one epoch. The latter could be especially promising for large language models, for which one epoch of training can already consume hours of training time.

\section{Acknowledgment}

We would like to thank Prof. Isabelle Guyon for participating with us in discussions that improved the quality of this work.

This material is based upon work supported by the U.S.\ Department of Energy 
(DOE), Office of Science, Office of Advanced Scientific Computing Research, under
Contract DE-AC02-06CH11357. This research used resources from the Argonne 
Leadership Computing Facility, which is a DOE Office of Science User Facility. 
This material is based upon work supported by ANR Chair of Artificial Intelligence HUMANIA ANR-19-CHIA-0022 and TAILOR EU Horizon 2020 grant 952215.
Felix Mohr participated through the project ING-312-2023 from Universidad de La Sabana, Campus del Puente del Común, Km. 7, Autopista Norte de Bogotá. Chía, Cundinamarca, Colombia.



\bibliographystyle{elsarticle-num} 
\bibliography{esann-short,tom,romain}



% % %% \bibitem{label}
% % %% Text of bibliographic item

% % \bibitem{}

% \end{thebibliography}

% \appendix

\end{document}